%% file: arxiv.tex
\documentclass[11pt]{article}

\usepackage{microsoft-tech-report}
\usepackage[numbers,sort&compress]{natbib}
\usepackage{microtype}

\input{math_commands.tex}

\input{results_macros.tex}

\input{human_review_macros.tex}
\input{preamble_common}

\tastepublictrue

\PassOptionsToPackage{hyphens}{url}
\usepackage{hyperref}
\usepackage{url}
\hypersetup{
  colorlinks=true,
  linkcolor=msftblue,
  citecolor=msftblue,
  urlcolor=msftblue,
  pdftitle={The Tasteful Agent: Measuring and Improving Taste in Long-Horizon Tasks},
  pdfauthor={Wenbo Pan, Zhichao Liu, Shujie Liu, Jingying Zeng, Chin-Yew Lin, Xianfeng Tang, Yan Lu, Qi He, Xiaohua Jia}
}

\techreportlabel{Preprint}
\techreportshorttitle{The Tasteful Agent}

\begin{document}
\thispagestyle{empty}

\noindent
\begin{minipage}[c]{0.5\linewidth}
\raggedright
{\footnotesize\msftsans\color{msftgray}\microsoftreportlabel}
\end{minipage}%
\begin{minipage}[c]{0.49\linewidth}
\raggedleft
{\msftdatefont\small\color{msftgray}September 2026}
\end{minipage}\par
\vspace{0.35em}
\noindent{\color{msftline}\rule{\linewidth}{0.8pt}\par}

\vspace{1.0em}
\begin{center}
{{\msfttitlefont\fontsize{21}{25}\selectfont\color{msftdark}
The Tasteful Agent: Measuring and Improving\\
Taste in Long-Horizon Tasks\par}}
\vspace{1.25em}

{\normalsize\rmfamily\color{msftdark}
Wenbo Pan$^{1,\dagger}$ \hspace{0.6em}
Zhichao Liu$^{2}$ \hspace{0.6em}
Shujie Liu$^{3}$ \hspace{0.6em}
Jingying Zeng$^{3}$ \hspace{0.6em}
Chin-Yew Lin$^{3}$\\[-0.1em]
Xianfeng Tang$^{3}$ \hspace{0.6em}
Yan Lu$^{3}$ \hspace{0.6em}
Qi He$^{3}$ \hspace{0.6em}
Xiaohua Jia$^{1}$\par
}
\vspace{0.22cm}

{\footnotesize\rmfamily\color{msftgray}
$^{1}$ City University of Hong Kong \quad
$^{2}$ Independent Researcher \quad
$^{3}$ Microsoft\par
}
\end{center}

\vspace{0.45em}
\begin{msfttitlebox}
\setlength{\parindent}{0cm}
\setlength{\parskip}{0.14cm}
\raggedright
\nohyphens

\begin{abstract}
\input{abstract}
\end{abstract}

\vspace{0.14cm}
{\setlength{\parskip}{0.06cm}\small
{{\color{msftblue}\githubicon}\hspace{0.35em}\msftmetalabel{Code}\href{\tasterepourl}{\tasterepourl}\par}
{\hficon\hspace{0.35em}\msftmetalabel{Dataset}\href{\tastedataurl}{\tastedataurl}\par}
{{\color{msftblue}\raisebox{-0.15ex}{\faEnvelope}}\hspace{0.35em}\msftmetalabel{Correspondence}
\href{mailto:wenbo.pan@my.cityu.edu.hk}{wenbo.pan@my.cityu.edu.hk}\par}
}
\vspace{0.08cm}
{\footnotesize\rmfamily\itshape\color{msftgray}
$^{\dagger}$ Work done during an internship at Microsoft.\par
}
\end{msfttitlebox}

\input{body}

\bibliographystyle{unsrtnat}
\bibliography{references}

\clearpage
\appendix
\input{appendix}

\end{document}

%% file: math_commands.tex
\usepackage{amsmath,amsfonts,bm}

\def\eqref#1{equation~\ref{#1}}

\def\1{\bm{1}}

\DeclareMathAlphabet{\mathsfit}{\encodingdefault}{\sfdefault}{m}{sl}
\SetMathAlphabet{\mathsfit}{bold}{\encodingdefault}{\sfdefault}{bx}{n}



%% file: results_macros.tex
\newcommand{\NumItems}{502}
\newcommand{\NumModels}{14}
\newcommand{\BestModel}{GPT-5.6 Sol}
\newcommand{\BestConsistent}{59.7\%}
\newcommand{\SecondModel}{GPT-5.5}
\newcommand{\SecondConsistent}{59.5\%}

%% file: human_review_macros.tex
\newcommand{\HumanReviewN}{100}
\newcommand{\HumanExplicitN}{172}
\newcommand{\HumanSupportedN}{170}
\newcommand{\HumanExplicitAgreement}{98.8\%}
\newcommand{\HumanDirectionalOverlap}{74}
\newcommand{\HumanDirectionalAgreementN}{73}
\newcommand{\HumanDirectionalAgreement}{98.6\%}
\newcommand{\HumanDirectionalKappa}{0.973}

%% file: preamble_common.tex
\usepackage{graphicx}
\usepackage{booktabs}
\usepackage{multirow}
\usepackage{xcolor}
\usepackage{amsmath}
\usepackage{amssymb}
\usepackage{enumitem}
\usepackage{wrapfig}
\usepackage{float}
\usepackage{tikz}
\usetikzlibrary{arrows.meta,positioning,calc,fit,backgrounds,decorations.pathreplacing}
\usepackage{varwidth}
\usepackage[most]{tcolorbox}
\graphicspath{{figures/}}

\definecolor{tastepurple}{HTML}{5B4BB7}
\definecolor{tasteteal}{HTML}{2A9D8F}
\definecolor{tastecoral}{HTML}{D96B62}
\definecolor{tasteblue}{HTML}{4A78A8}
\definecolor{tasteorange}{HTML}{D88A3D}
\definecolor{tasteink}{HTML}{263238}
\definecolor{tastegray}{HTML}{7B8794}
\input{figures/system/taste_figure_style.tex}
\newcommand{\taste}{\textsc{Taste-Bench}}
\newif\iftastepublic
\tastepublicfalse
\newcommand{\tasterepourl}{https://github.com/wbopan/tastebench}
\newcommand{\tastedataurl}{https://huggingface.co/datasets/wenbopan/taste-bench}
\usepackage{fontawesome5}
\newcommand{\githubicon}{\raisebox{-0.15ex}{\faGithub}}
\newcommand{\hficon}{\raisebox{-0.3ex}{\includegraphics[height=1.1em]{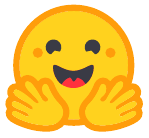}}}

\definecolor{remarkborder}{HTML}{5F83BF}  
\definecolor{remarktitlebg}{HTML}{2B4A7A} 
\definecolor{remarkbg}{HTML}{F2F6FB}      

\NewDocumentEnvironment{remark}{O{} O{\linewidth}}{%
  \begin{tcolorbox}[
    width=#2,
    enhanced,
    colback=remarkbg,
    colframe=remarkborder,
    boxrule=0.8pt,
    arc=2pt,
    left=7pt,
    right=7pt,
    bottom=4pt,
    top=4pt,
    before skip=5pt,
    after skip=5pt,
    title={#1},
    fonttitle=\bfseries\sffamily\footnotesize,
    coltitle=white,
    varwidth boxed title*=-2mm,
    boxed title style={
      colback=remarktitlebg,
      colframe=remarktitlebg,
      boxrule=0pt,
      arc=2pt,
      left=5pt,
      right=5pt,
      top=2pt,
      bottom=2pt
    },
    attach boxed title to top left={
      xshift=7pt,
      yshift*=-\tcboxedtitleheight/2
    }
  ]
  \small
}{%
  \end{tcolorbox}
}

%% file: figures/system/taste_figure_style.tex
\tikzset{
  taste flow/.style={-{Stealth[length=2.2mm,width=1.5mm]},draw=tasteink,line width=0.75pt},
  taste softflow/.style={-{Stealth[length=2mm,width=1.35mm]},draw=tastegray,line width=0.65pt},
  taste box/.style={draw=tasteink,line width=0.65pt,rounded corners=1.4pt,
    align=center,inner xsep=5pt,inner ysep=4pt,font=\sffamily\scriptsize,text=tasteink},
  taste visible/.style={taste box,fill=tasteblue!11,draw=tasteblue!90!black},
  taste candidate/.style={taste box,fill=tasteorange!15,draw=tasteorange!95!black},
  taste future/.style={taste box,fill=black!5,draw=tastegray,dashed},
  taste measure/.style={taste box,fill=tastepurple!10,draw=tastepurple},
  taste result/.style={taste box,fill=tasteteal!11,draw=tasteteal!90!black},
  taste warn/.style={taste box,fill=tastecoral!10,draw=tastecoral!95!black},
  taste title/.style={font=\sffamily\scriptsize\bfseries,text=tasteink,align=center},
  taste note/.style={font=\sffamily\tiny,text=tastegray,align=center},
  taste tag/.style={font=\sffamily\tiny\bfseries,text=white,rounded corners=1pt,
    inner xsep=3pt,inner ysep=1.5pt}
}

%% file: abstract.tex
LLM agents increasingly work on long-horizon tasks, and the decisions they make along the way, such as which hypothesis to test or which implementation to build on, determine the outcome of the whole run.
Making these decisions well is becoming a key capability for both engineering and research agents.
We refer to the ability to make good long-horizon decisions as the \emph{taste} of an agent.
While existing benchmarks measure the end-to-end success of agents on long-horizon tasks, none of them measures the taste of an agent.
To address this problem, we build \taste{}, a benchmark of taste questions constructed automatically from trajectories that agents produced in engineering and research tasks.
Each question presents a \emph{decision fork}, a point in a trajectory where multiple directions are available and one of them leads to a better outcome, and the evaluated model chooses among these directions without seeing what happens after the fork.
We mine these forks automatically from parallel attempts at the same task and from detours inside a single trajectory, without needing human annotation.
We evaluate frontier models on \taste{} and find that the best model answers only \BestConsistent{} of the questions correctly.
We further find that forks whose deciding evidence appears later in the trajectory are much harder for every model, and that a larger reasoning budget does not improve the accuracy.
Finally, we show that taste can be trained.
We distill the judgment of a teacher that has seen the outcome into a student model, and the student makes better decisions on unseen tasks and improves end-to-end success on held-out SWE-bench Pro tasks.

%% file: body.tex
\section{Introduction}

LLM agents increasingly work on long-horizon tasks, and the length of the tasks they can complete keeps increasing \citep{kwa2025longtasks,metr2026timehorizon}.
For instance, recent systems conduct machine-learning research from idea to paper \citep{lu2026automation,mitchener2025kosmos}, evolve large software projects across releases \citep{thai2025sweevo}, and refine their own scaffolds during deployment \citep{xia2025liveswe}.
In these tasks, the agent makes many decisions whose influence is not limited to the current step, such as which hypothesis to test, which implementation to build on, or which experiment to run next.
Making these decisions well is becoming a key capability for agents \citep{novikov2025alphaevolve,zhang2025darwingodel}.
However, a wrong decision often looks reasonable at the moment, and its cost appears only much later, after the agent has spent a large part of its budget.

We refer to the ability to make good long-horizon decisions as the \emph{taste} of an agent.
While previous research has measured the end-to-end performance of agents on long-horizon tasks \citep{liu2023agentbench,jimenez2023swebench,deng2025swebenchpro,huang2023mlagentbench,wijk2024rebench}, these benchmarks only report whether the agent finishes the task and provide no measure of the quality of the decisions made along the way.
However, measuring these decisions directly is difficult, because the result of a long-horizon decision is not immediately visible, so a good choice and a bad choice can look equally reasonable at the decision point.
Judging decision quality also requires deep domain expertise, so using human annotation is expensive and hard to scale across new domains.
This work asks whether the taste of an agent can be measured automatically, without expert annotation.

\begin{wrapfigure}{R}{0.5\textwidth}
    \centering
    \vspace{-10pt}
    \includegraphics[width=\linewidth]{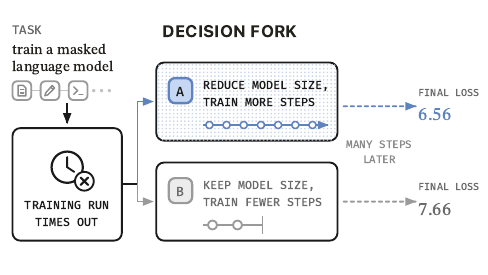}
    \vspace{-31pt} 
    \caption{\textbf{A decision fork from a machine-learning trajectory.} The evaluated model chooses before the later losses reveal that candidate A is the better direction.}
    \label{fig:fork-example}
    \vspace{-8pt}
\end{wrapfigure}

Our key observation is that the later part of a trajectory provides hindsight evidence for its earlier decisions.
In practice, agent systems routinely make several attempts at the same task, and the attempts often diverge into different directions in the middle of the run.
When this happens, the recorded outcome of each attempt identifies the better direction, so the trajectories themselves provide a labeled comparison.
We call such a divergence point a \emph{decision fork}.
To measure taste automatically, we freeze the trajectory at the fork, hide all later work, and ask a model to choose between the two directions.
Figure~\ref{fig:fork-example} shows one such fork from a machine-learning trajectory.

We collect decision forks from both software engineering and machine-learning research trajectories, and we build \taste{}, a benchmark of \NumItems{} taste questions.
Each question contains the original task, the trajectory prefix up to the fork, and the two directions at the fork, where the hidden later work identifies the better one.
The forks are mined in two ways, from parallel trajectories, where independent attempts at the same task diverge, and from detour trajectories, where an agent corrects itself inside a single run, and Section~\ref{sec:benchmark} describes the two constructions.
Because the supervision is mined from work that agents already performed, the benchmark scales with the trajectories that agent systems produce.
We further assess the mined labels through a human review, which shows close agreement with the labels when reviewers identify a better direction after seeing the recorded outcomes.

Our core contributions are summarized as follows:
\begin{itemize}[leftmargin=15pt,itemsep=1pt,topsep=3pt]
    \item We formalize the taste of an agent as its ability to choose the better direction at a decision fork, and we show that this ability can be measured from hindsight over existing trajectories.
    \item We build and release \taste{},\iftastepublic\footnote{\hficon~Dataset: \url{\tastedataurl}; \githubicon~Code: \url{\tasterepourl}.}\fi{} a benchmark of such taste questions covering software engineering and machine-learning research, and we show that questions become harder as the time horizon of the fork increases.
    \item We show that taste is trainable.
    Distilling the reasoning of a teacher given the correct direction improves judgments on unseen tasks and produces end-to-end gains on held-out agent tasks.
\end{itemize}

\section{Measuring Taste via Long-Horizon Judgment}
\label{sec:measurement}

We view taste as a form of \emph{long-horizon judgment}, where the outcome of the better direction is not visible at decision time and appears only in the later work.
For instance, a clean implementation passes the same tests as a hasty one at the time of writing, and its advantage appears only when every later change becomes easier.
Under this view, measuring taste means testing whether the direction a model chooses at decision time is the one whose advantage appears in the later work.

To run this test without expert annotation, we mine hindsight from existing agent trajectories.
A trajectory records both the judgments made along the way and the later work after them, so we can use the outcome of a trajectory to estimate the quality of its judgments.

However, the outcome of a single trajectory does not directly show whether a judgment was good or bad.
This is because, besides the judgment, execution quality and the environment also determine the outcome.
To separate the effect of the judgment from these factors, we look for comparisons across trajectories in which everything except the judgment is the same.
Such comparisons are common when an agent tries the same task multiple times, because the attempts sometimes share an equivalent prefix up to some point and then diverge into different judgments at that point.
Such a point is a decision fork, and we refer to the parts of the attempts after the fork as its \emph{branches}.
Because the attempts share an equivalent prefix before the fork and sampling randomness assigns the judgments to the branches, the outcomes of the branches differ mainly because of the judgments at the fork.
The later work on each branch also affects its outcome, so we keep only the forks whose outcomes clearly follow from the judgments, as Section~\ref{sec:benchmark} describes.

Formally, every decision fork defines one taste question.
Consider attempts at a task $q$ that share the trajectory prefix $h_t=(o_0,a_0,\ldots,o_t)$, where $o$ denotes an observation and $a$ an action, and diverge into two candidate directions $c_1$ and $c_2$ at the fork time $t$.
Each branch $i$ then runs to completion and produces evidence $E_i$, such as test results or research scores, and an outcome measure $U$ maps this evidence to a scalar.
Under this setup, a model with good taste picks the candidate with the higher expected outcome $\mathbb{E}[U\mid h_t,c_i]$ before any evidence is available.
The realized outcomes therefore estimate these expectations under identical conditions, so we label the fork with the \emph{supported candidate}, the candidate whose branch produces the better outcome,
\begin{equation}
    y=\arg\max_{i\in\{1,2\}} U(E_i).
\end{equation}
The question is then $x=(q,h_t,c_1,c_2)$, where everything after the fork time $t$ is hidden from the evaluated model.
The evaluated model $\pi$ receives $x$ and returns one choice $\pi(x)$, and over a set of questions we estimate its taste $\widehat{T}(\pi)$ as the fraction of questions on which $\pi(x)=y$.

\section{Taste-Bench}
\label{sec:benchmark}

In this section, we apply the hindsight measurement of Section~\ref{sec:measurement} to real agent trajectories and present \taste{}, a benchmark of \NumItems{} taste questions.

Turning a raw trajectory into a question means recovering the task $q$, the trajectory prefix $h_t$, the candidate directions, and the label $y$ from an unstructured record.
We recover them through two constructions, one from \emph{parallel trajectories} and one from \emph{detour trajectories}.
Moreover, the two constructions are complementary, because parallel trajectories contain wrong judgments that the agent never notices, while detour trajectories contain wrong judgments that the agent corrects later.

\subsection{Forks from parallel trajectories}

In the first construction, we use parallel trajectories, exactly the situation that Section~\ref{sec:measurement} describes, where attempts at the same task diverge at the same fork and the realized outcomes identify the supported candidate.
In such attempts, the agent does not notice the wrong direction, and the attempt still runs to completion.
To construct one question, we align a pair of attempts with opposite outcomes at the fork where they diverge.
The part before the fork is equivalent on both sides and becomes the prefix, and the two directions at the fork, each written as a short neutral description, become the candidates.
The recorded outcome of each attempt then determines the supported candidate, so the branch that passes the tests or achieves the experimental objective provides the label.
This construction captures the wrong directions that an agent never notices, and the second construction captures the wrong directions that an agent takes and corrects inside a single trajectory.

\subsection{Forks from detour trajectories}

In the second construction, we use detour trajectories, where an agent corrects itself inside a single run, and the abandoned direction and the later recovery form the two candidates.
Here, the record itself marks the mistake, because the agent abandons the direction.
To identify a detour, a generator model reads one complete trajectory together with its outcome, and it locates three events in order: the step where the agent takes a direction, the observed failure that ends this direction, and the step where the agent takes a different direction that completes the task.
The generator then writes the two directions as two candidates in parallel wording, and it rejects the trajectory when the better direction is only nameable after seeing the failure or when either direction is not a plausible choice at the fork (Appendix~\ref{app:prompts}).
To construct one question, we place the fork at the step right before the agent takes the abandoned direction.
The part before the fork becomes the prefix, which both candidates then extend, and the generator checks that the prefix does not reveal the failure or the later fix.
The recorded outcome again determines the supported candidate, because the abandoned direction produces an observed failure and the recovery produces task completion.
The question therefore tests whether the evaluated model can recognize the failure earlier than the acting agent did.

\begin{figure*}[t]
    \centering
    \includegraphics[width=0.82\textwidth]{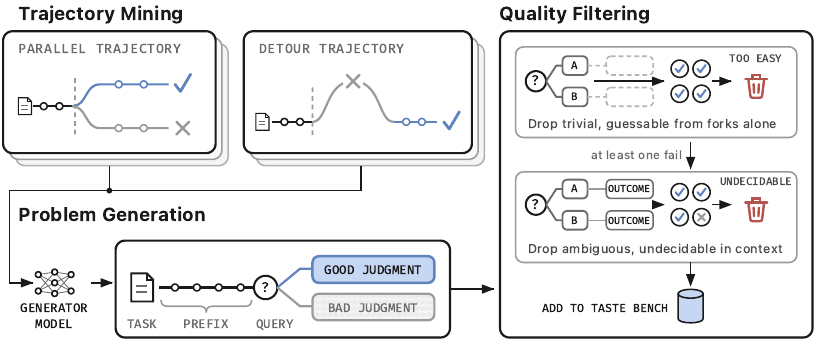}
    \vspace{-3pt}
    \caption{\textbf{Construction and filtering of \taste{}.} The parallel construction contrasts branches of the same task, while the detour construction contrasts an abandoned direction with the later recovery. The filtering stage removes trivial questions, whose answer can be guessed from the wording of the candidates alone, and undecidable questions, whose label is not clearly consistent with the full record of the task.}
    \label{fig:construction}
\end{figure*}

\subsection{Mining, filtering, and composition}
\label{sec:mining}

We mine both kinds of forks from trajectory pools in two domains, and we keep only the forks that form valid questions.
Figure~\ref{fig:construction} summarizes the pipeline.

\noindent\textbf{Mining.} The trajectory pools cover two domains.
The engineering pool contains 2,677 graded rollouts that we collect by running GPT-5.4 and GPT-5.5 agents on 517 SWE-bench Pro tasks \citep{deng2025swebenchpro}.
The research pool contains 1,132 agent runs on 47 AI R\&D tasks from RE-Bench and the research subset of HCAST \citep{wijk2024rebench,rein2025hcast}, and we download these runs from MALT, the public transcript release of METR \citep{parikh2025malt}.\footnote{\url{https://huggingface.co/datasets/metr-evals/malt-public}}
A generator model reads these trajectories and proposes \emph{candidate forks} of the two kinds above.
The generator applies a rubric and keeps only the candidate forks from which a valid question can be extracted, and Appendix~\ref{app:forks} lists the checks of this rubric.

\noindent\textbf{Filtering.} A candidate fork that passes the rubric can still fail as a question in two ways.
\textbf{(1) Trivial.} The answer can be guessed from the wording of the two candidates alone, so the question does not test judgment over the trajectory.
\textbf{(2) Undecidable.} The recorded outcome is not clearly consistent with the label.
We remove both kinds with a set of judge models that does not include the generator.
Each judge first answers the question given only the two candidates without the trajectory, and a question is discarded as trivial when every judge answers it correctly.
Each judge then reads the full record of the task, the trajectory, and the outcome, and a question is included in the release only when every judge agrees with its label.
In this way, every released question is answered incorrectly by at least one judge when the trajectory is hidden, and its label is confirmed by every judge when the full record of the task is visible.

\noindent\textbf{Composition.} The generator proposes 4,657 candidate forks from the two pools, and 10.8\% of them pass all filters.
We organize the resulting \NumItems{} questions in a $2\times2$ design that crosses the construction (parallel or detour) with the domain (engineering or research), where engineering provides 390 questions and research provides 112.
Appendix~\ref{app:counts} gives the counts at each stage per cell.

\noindent\textbf{Human review.} We assess the quality of the mined labels through a human review of \HumanReviewN{} sampled questions, where two reviewers separately choose a direction for each question, then see summaries of the recorded continuations and outcomes and judge which decision was better.
We exclude judgments without a clear A/B preference from the agreement analysis.
Of the \HumanExplicitN{} explicit A/B judgments, \HumanSupportedN{} agree with the mined label, yielding \textbf{\HumanExplicitAgreement{} agreement}, and on the \HumanDirectionalOverlap{} questions where both reviewers choose A or B, they agree with each other on \HumanDirectionalAgreement{}, with Cohen's $\kappa=\HumanDirectionalKappa{}$, and Appendix~\ref{app:human-review} reports the results for each reviewer.

\subsection{Evaluation protocol}
\label{sec:protocol}

The questions of Section~\ref{sec:mining} are instances of the question $x$ defined in Section~\ref{sec:measurement}, so we present each question to the evaluated model, and the answers over the release estimate the taste $\widehat{T}(\pi)$.

However, two-choice questions suffer from position bias \citep{shi2024judges}, and merely swapping the order of the two candidates changes the answers of many models.
To address this, each question is evaluated twice, once in a deterministic seeded order and once in the exact reverse order.
Throughout the paper, \emph{accuracy} refers to this estimate, where a question counts as correct only when both orders are answered correctly, so an answer that flips with the order does not count as taste.
We also report the \emph{mean accuracy} over the two orders.

\section{Experimental Results}
\label{sec:results}

\subsection{Setup}

\noindent\textbf{Models.} To measure the taste of current LLM agents, we evaluate \NumModels{} contemporary models on the \NumItems{} questions of \taste{}.
The models cover the major frontier families, including Claude \citep{anthropic2026opus5,anthropic2026sonnet5}, GPT \citep{openai2026gpt54mini,openai2026gpt55,openai2026gpt56}, Grok \citep{xai2026grok420,xai2026grok45}, DeepSeek \citep{deepseek2026v4}, GLM \citep{zhipu2026glm5,zai2026glm52}, MiniMax \citep{minimax2026m3}, and Mistral \citep{mistral2026medium}, and they all answer under the same interface and token budget.

\noindent\textbf{Reporting.} The primary metric is the accuracy defined in Section~\ref{sec:protocol}, and headline numbers report the Average, the 1:1 mean of the research and engineering subset accuracies.
Because a question counts as correct only when both orders are answered correctly, the accuracy of random guessing is 25\%, the product of one half per order, and a model that always prefers the same position scores 0\%.

\subsection{Accuracy of current models}

\begin{figure*}[t]
    \centering
    \includegraphics[width=0.85\textwidth]{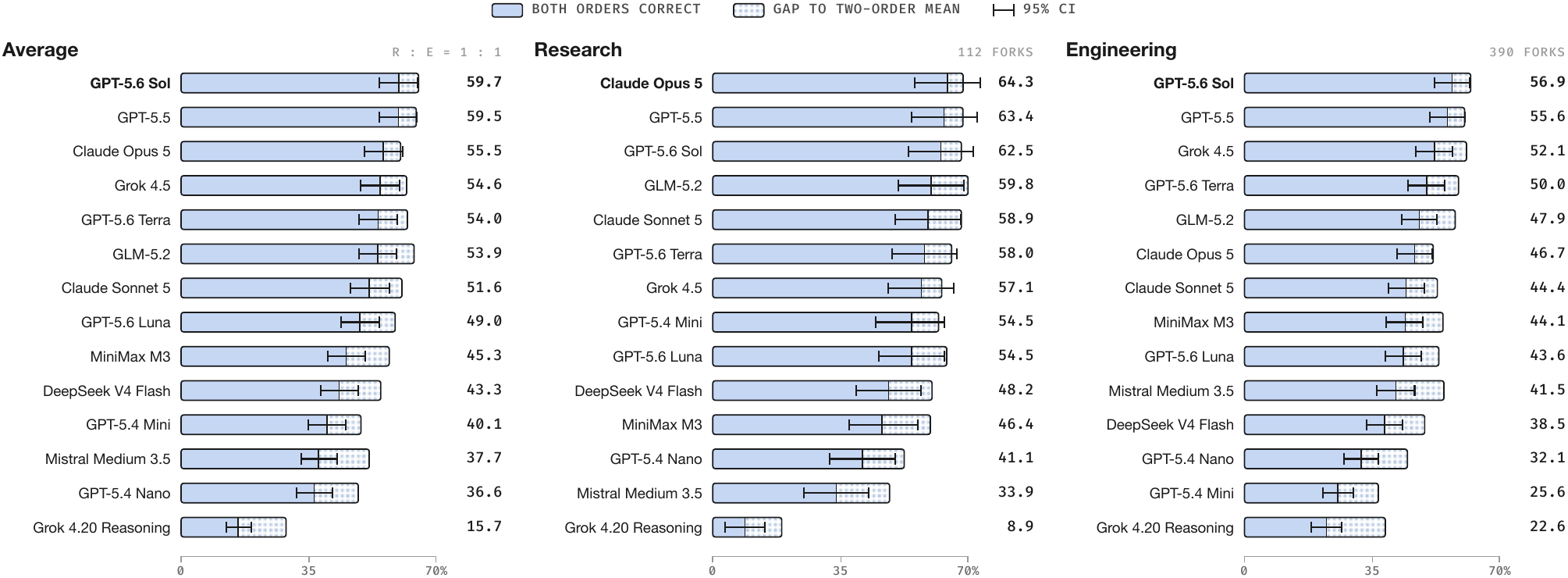}
    \vspace{-5pt}
    \caption{\textbf{Model performance on \taste{}.} The first column reports the Average defined in Section~\ref{sec:results}, followed by the research and engineering subsets. Each column shows the model names again and is sorted independently from highest to lowest within that split. Solid bars show accuracy when both candidate orders are answered correctly. The dotted extension shows the mean accuracy over the two orders. Whiskers show 95\% item-bootstrap intervals for the solid-bar endpoint, and the printed values report that endpoint. Unparseable responses count as incorrect.}
    \label{fig:ranking}
\end{figure*}

As Figure~\ref{fig:ranking} shows, the models differ widely on \taste{}, and none of them is close to solving it.
Specifically, \BestModel{} is the best model, with an Average accuracy of \BestConsistent{}, and \SecondModel{} is close behind at \SecondConsistent{}, while the accuracies of the remaining models are widely dispersed below them, with additional variation between the research and engineering subsets.

\begin{remark}[Finding 1]
\textbf{Current frontier models show limited taste.} Even the strongest models cannot reliably identify the better direction at decision time, although every question is a binary choice.
\end{remark}

Across the four cells of the release, detour forks are harder than parallel forks in both domains, and this gap exceeds the gap between the two domains (Appendix~\ref{app:full-results}).

\subsection{Effect of the time horizon}

\begin{figure*}[t]
    \centering
    \includegraphics[width=0.88\textwidth]{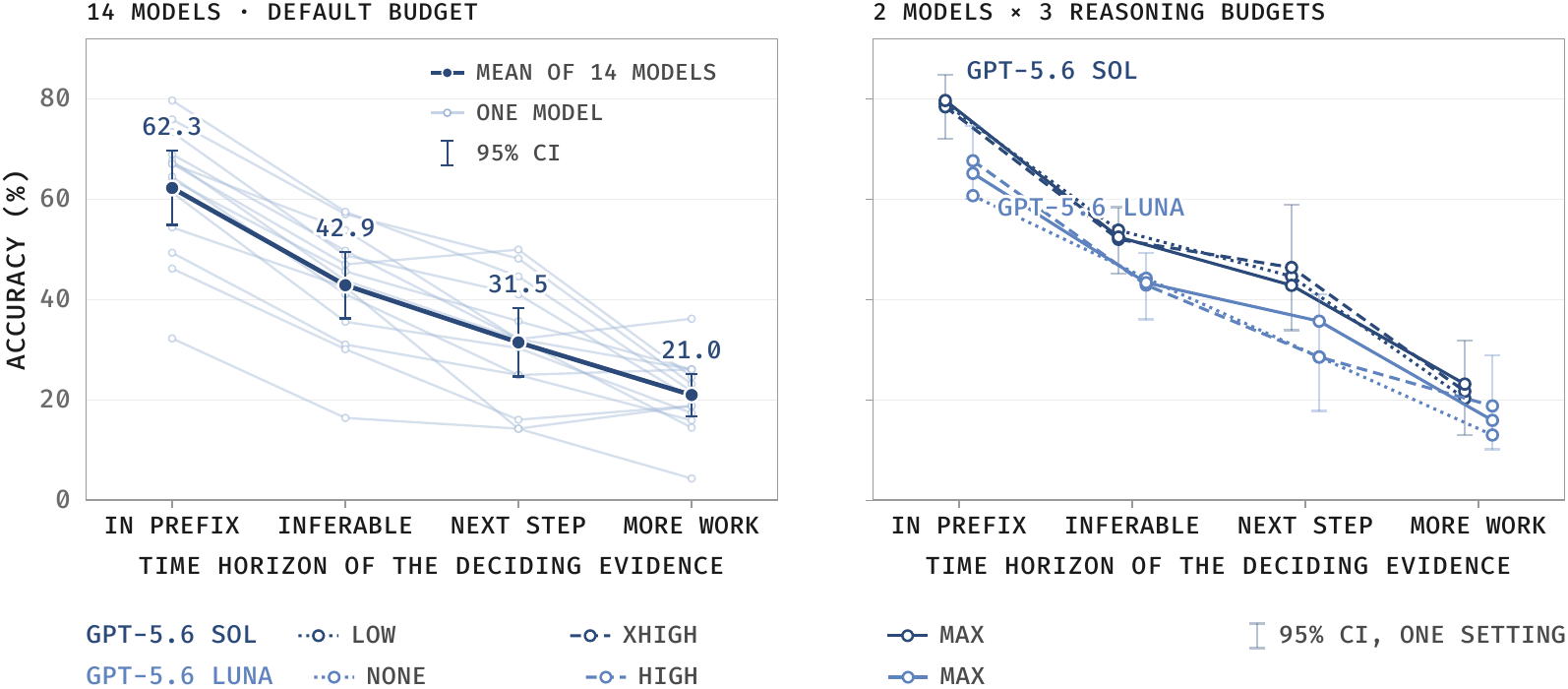}
    \vspace{-4pt}
    \caption{\textbf{Accuracy by time horizon and by reasoning budget.} Each fork is annotated with a time horizon on four levels, from in prefix to more work. \textbf{Left.} The dark line is the mean over the \NumModels{} models, with 95\% $t$-intervals of that mean, and the faint lines show the individual models. \textbf{Right.} Accuracy at the same levels under three reasoning-effort settings per model, where the whisker is the 95\% bootstrap interval of one setting of one model.}
    \label{fig:seh}
\end{figure*}

Forks whose deciding evidence appears late in the trajectory should be harder.
To turn this intuition into a measurable quantity, we annotate every fork with a \emph{time horizon}, which is how far into the future an observer at the fork must see before the supported candidate is clearly justified.
A judge model reads the task, the prefix, the two candidates, and the supported label, and it assigns one of four ordinal levels according to which part of the record justifies the supported candidate: \textbf{in prefix}, where a decisive fact that excludes one candidate is already visible in the prefix, \textbf{inferable}, where no single decisive fact is visible but the hints in the prefix together justify the supported candidate, \textbf{next step}, where the first observation after the fork justifies it, and \textbf{more work}, where the justification requires a completed local check or more substantial later work.
Appendix~\ref{app:horizon} gives the judge prompt and the number of questions per level.

The left panel of Figure~\ref{fig:seh} plots accuracy over the four levels, and accuracy falls on average across models as the time horizon increases.
Specifically, the mean over the \NumModels{} models falls from 62.3\% at the in-prefix level to 21.0\% at the more-work level, near the 25\% score of random guessing.

\begin{remark}[Finding 2]
\textbf{Model errors concentrate on forks with a long time horizon.} Accuracy falls on average as the horizon increases, so answering the questions requires predicting the later work.
\end{remark}

\subsection{Effect of the reasoning budget}
\label{sec:effort}

However, the falling accuracy of Figure~\ref{fig:seh} may be an effect of the reasoning budget, because a model that reasons longer at the fork may predict more of the later work.
To test this, we rerun \taste{} under three reasoning-effort settings for two models, and we keep the questions, the prompt, the token limit, and the protocol of Section~\ref{sec:protocol} unchanged, which produces six conditions and 6,024 responses in total across the two models.

The additional reasoning does not change the accuracy.
Specifically, moving from the lowest to the highest reasoning-effort setting changes the accuracy of GPT-5.6 Sol by $-0.2$ points and that of GPT-5.6 Luna by $+2.2$ points.
The right panel of Figure~\ref{fig:seh} repeats the comparison at every time horizon, where the settings of each model overlap at every level.

We also record where the models spend this budget.
At every setting with reasoning enabled, the two models produce the most reasoning tokens at the more-work level, which is also the level with the lowest accuracy (Appendix~\ref{app:effort}).
This suggests that the models recognize the hard forks and reason longest on them, and that the deciding evidence at these forks appears only in the later work.

\begin{remark}[Finding 3]
\textbf{A larger reasoning budget does not improve taste.} The accuracy is unchanged at every time horizon, and the models reason longest at the level where their accuracy is lowest, which suggests that the deciding evidence at these forks appears only in the later work.
\end{remark}

\subsection{Comparison with end-to-end benchmarks}
\label{sec:vs-swebench}

\begin{wrapfigure}{r}{0.38\textwidth}
    \centering
    \vspace{-18pt}
    \includegraphics[width=\linewidth]{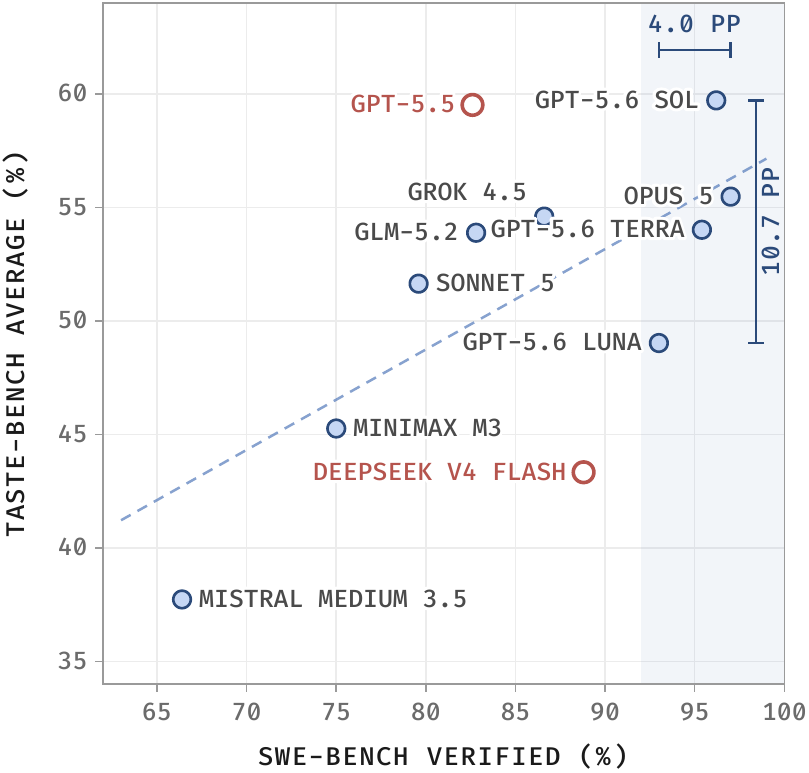}
    \vspace{-20pt}
    \caption{\textbf{\taste{} against SWE-bench Verified.} Each point is one model. The dashed line is the least-squares fit, red points deviate most from it, and the band marks the four highest SWE-bench Verified scores.}
    \label{fig:vs-swebench}
    \vspace{-10pt}
\end{wrapfigure}

Finally, we compare the ranking of Figure~\ref{fig:ranking} against an end-to-end agent benchmark, because the taste measurement is useful only when it is not a restatement of end-to-end ability.
We compare the Average of each model with its public score on SWE-bench Verified \citep{jimenez2023swebench,chowdhury2024sweverified}, and we take the public scores from the Vals AI leaderboard \citep{valsai2026swebench}, which reports every model of Figure~\ref{fig:ranking} under one shared harness.
We exclude three models whose responses are unparsable on more than 9\% of the presentations, so 11 models remain in the comparison.

Figure~\ref{fig:vs-swebench} shows the comparison, and the two benchmarks are only partly correlated.
Specifically, the Pearson correlation between the Average and SWE-bench Verified is $r=+0.63$, so SWE-bench Verified explains $R^2=0.39$ of the variance between the models.
However, the correlation on the engineering subset is only $r=+0.37$, although this subset is mined from SWE-bench Pro tasks and is therefore the closest comparison.
Moreover, the models at the top of SWE-bench Verified are clearly separated on \taste{}.
The four highest models on SWE-bench Verified are within 4.0 points of each other there, while their Averages are 10.7 points apart (Appendix~\ref{app:vs-swebench}).

\section{Generalizing Taste through Distillation}
\label{sec:distill}

\begin{figure}[!ht]
    \centering
    \includegraphics[width=0.85\textwidth]{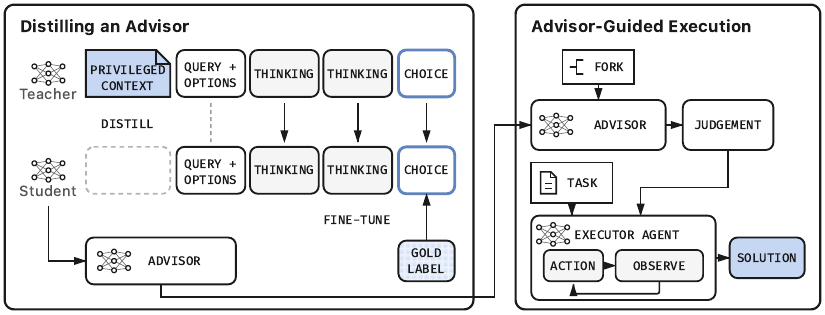}
    \caption{\textbf{Overview of distillation and advice injection.} \textbf{Left.} We distill the reasoning of a teacher that has seen the supported candidate into a student that sees only the question. \textbf{Right.} At evaluation time, the student's judgment at each fork is written as advice into the task context, and a fixed executor completes the task independently.}
    \label{fig:distill-overview}
\end{figure}

The results above show that current models judge these forks poorly.
We therefore ask whether we can also train taste from the trajectories that measure it.
The questions of Section~\ref{sec:benchmark} already contain the needed supervision, because each question contains two candidate directions together with the outcome that determines the supported candidate.
In this section, we distill this judgment into the weights of a model and test whether it transfers to unseen tasks and improves end-to-end task success (Figure~\ref{fig:distill-overview}).
Throughout this section, the base model is Qwen3.6-27B, and training updates LoRA adapters \citep{hu2022lora}.

\subsection{Distillation recipe}
\label{sec:recipe}

\noindent\textbf{Task-disjoint folds.} The training pool contains the 390 engineering questions of Section~\ref{sec:benchmark}, split into two task-disjoint folds.
Each student trains on one fold and is evaluated on the other fold, so no student sees the source task of any evaluation question during training.

\noindent\textbf{Distilling reasoning instead of fitting labels.} Fine-tuning directly on the supported labels is the obvious objective, but each question provides a single binary label and no intermediate reasoning, so a model fitted on the labels memorizes the answers without learning the judgment.
We therefore distill outputs that contain a complete reasoning sequence, which a privileged teacher produces.
The teacher and the student are the same frozen base model with different contexts.
The teacher context contains the question together with a \emph{demonstration}, a short description of the supported candidate.
Because the teacher already knows the answer, it reliably selects the supported candidate in its freely generated reasoning.
Training then aligns the same teacher-generated tokens under the student context, which contains only what the benchmark shows an evaluated model, namely the task, the trajectory prefix, and the two shuffled candidates.
The loss is an SDPO-style token-level distillation with a forward KL \citep{huebotter2026sdpo}, computed over the reasoning tokens and the final choice.
We compute this KL on continuations sampled from the teacher rather than from the student, and Appendix~\ref{app:recipe} explains this choice and the leakage control of the recipe.

\subsection{Transfer to unseen tasks}
\label{sec:transfer}

\begin{figure}[t]
    \centering
    \begin{minipage}[c]{0.47\textwidth}
        \centering
        \includegraphics[width=\linewidth]{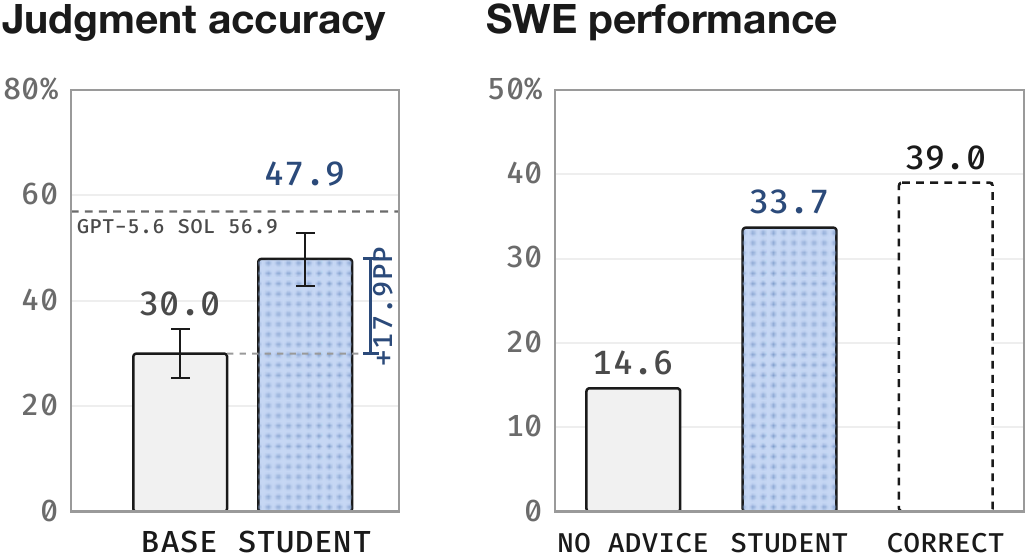}
    \end{minipage}\hfill
    \begin{minipage}[c]{0.49\textwidth}
        \caption{\textbf{Distillation results.} \textbf{Left.} On the 390 held-out engineering questions, scored on both orders as in Figure~\ref{fig:ranking}, the distilled student reaches a higher accuracy than the base model. Whiskers are 95\% item-bootstrap intervals, and the dashed line marks the accuracy of GPT-5.6 Sol on the same questions (56.9\%). \textbf{Right.} On held-out SWE-bench Pro tasks, the executor's success rate with student advice reaches most of the upper bound, the dashed bar, where the advice at every fork is correct.}
        \label{fig:distill-results}
    \end{minipage}
\end{figure}
On the training fold, distillation raises the fraction of questions that the student answers correctly in a single order from 48.6\% to 92.9\%, but this gain does not show whether the student learns the judgment or memorizes the answers, and the task-disjoint evaluation separates the two (Figure~\ref{fig:distill-results}, left).
On the held-out fold, we score the student and the base model with the protocol of Section~\ref{sec:protocol}.
Under this protocol, the student reaches an accuracy of 47.9\% and the base model 30.0\%, while the mean accuracy over the two orders rises from 42.7\% to 62.4\%.
The student therefore gains 17.9 percentage points on questions from tasks that it never saw in training, so the judgment transfers to unseen tasks.
Appendix~\ref{app:transfer} places the student in the engineering ranking of Figure~\ref{fig:ranking} and gives the rule that combines the two orders into one decision per question, which Section~\ref{sec:e2e-results} needs.

\subsection{End-to-end evaluation}
\label{sec:e2e-results}

We now let the student advise an agent and measure task success on held-out SWE-bench Pro tasks.

For each held-out task, we write the judgment at each fork as one note of \emph{advice}, which contains the situation at that fork, the candidate to avoid, and the candidate to take.
The pipeline of Section~\ref{sec:benchmark} mined these forks from earlier runs on these tasks, so the advice is available before the new run starts.
We then place advice into the task context, and a fixed Qwen3.6-27B executor agent completes the task independently of the student.

We evaluate three settings on the same 41 held-out SWE-bench Pro tasks: \textbf{no advice}, \textbf{correct advice}, where the advice at every fork specifies the supported candidate, and \textbf{student advice}, where the distilled student answers the taste question of every fork, and the advice recommends the candidate that the student selects.
The executor completes the tasks under each setting, and we measure its success rate using the official SWE-bench Pro evaluation.

Figure~\ref{fig:distill-results} (right) summarizes the outcomes. With correct advice, the success rate rises from 14.6\% to 39.0\%, a gain of 24.4 percentage points.
Because the advice at every fork is correct in this setting, this gain is the upper bound of what the advice can provide.

The student achieves most of this bound: the executor reaches a success rate of 33.7\% with student advice, a gain of 19.1 percentage points over no advice.
Appendix~\ref{app:e2e} gives the executor configuration and advice format.
These 41 tasks are unseen during training, so the better judgment of the student also improves task success outside its training distribution.

\begin{remark}[Finding 4]
\textbf{Taste can be distilled, and better judgment produces gains in task success.} The student makes better judgments on unseen tasks, and its advice raises the executor's success rate from 14.6\% to 33.7\%.
\end{remark}

\section{Related Work}

\paragraph{Long-horizon agent evaluation.}
AgentBench evaluates multi-step interaction \citep{liu2023agentbench}, SWE-bench and SWE-bench Pro evaluate software agents on repository tasks \citep{jimenez2023swebench,deng2025swebenchpro}, and MLAgentBench and RE-Bench extend evaluation to ML experimentation and AI R\&D \citep{huang2023mlagentbench,wijk2024rebench}.
These benchmarks test whether an agent completes a long task, whereas \taste{} tests whether the agent chooses the better direction inside the task.

\paragraph{Judging research directions before the outcome.}
Recent work asks whether a model can judge a research direction before its outcome is known, by predicting which of two ideas performs better \citep{wen2025predicting}, preferring one of two ML solutions before executing them \citep{zheng2026predict}, or choosing the better of two AI-safety research proposals against expert ratings \citep{baig2026taste}, and judgments of ideas before execution often flip after execution \citep{si2025ideation}.
These works judge standalone ideas or solutions, whereas \taste{} judges forks inside executed trajectories, so each question carries the situation of the agent and the recorded outcome labels it.

\paragraph{Process evaluation and step-level judgment.}
Process supervision scores steps with human labels \citep{lightman2023verify} or with the outcomes of rollouts from each step \citep{wang2024mathshepherd}.
For agents, preferences between explored and expert trajectories \citep{song2024trial} and between verified alternatives at critical steps \citep{li2026cso} have trained policies, and a process reward model scores candidate actions at test time \citep{chae2025webshepherd}.
Agent-as-a-Judge and AgentRewardBench judge whole trajectories \citep{zhuge2024agentasajudge,lu2025agentrewardbench}, and Who\&When and AgenTracer attribute a failure to its decisive step, which models localize poorly \citep{zhang2025whichagent,zhang2026agentracer}.
\taste{} instead labels the better direction with the outcome the trajectory later records, and the evaluated model chooses before seeing it.

\paragraph{Distillation and internalizing context.}
Knowledge distillation trains a student to match a teacher's output distribution \citep{hinton2015distilling}, and context distillation uses the same model conditioned on extra context as the teacher \citep{askell2021general,snell2022learning}, as do STaR with the correct answer \citep{zelikman2022star}, LEAP with privileged state \citep{choudhury2025leap}, and SDPO with environment feedback \citep{huebotter2026sdpo}.
Our recipe follows SDPO but replaces the feedback with a demonstration of the supported candidate and samples the targets from the teacher rather than from the student \citep{agarwal2024gkd}, so it transfers in-context judgment into the weights.
Advisor models similarly steer an executor with advice from a small trained model \citep{asawa2026advisor}.

\section{Conclusion}

In this work, we study taste, the ability of an agent to choose the better direction before the outcome is visible.
We show that this ability can be measured from existing trajectories, and we build \taste{}, a benchmark of \NumItems{} taste questions labeled by recorded outcomes.
Distilling the reasoning of a teacher given the supported candidate transfers judgment to questions from unseen tasks, and injecting the student's judgment as advice improves end-to-end success.
We hope \taste{} provides a practical basis for measuring and training the judgment of long-horizon agents.

\ifdefined\ificlrfinal
  \pagebreak
\fi

\section*{Reproducibility Statement}

All reported results use one fixed version of the benchmark, and the appendix specifies the constructions, the prompts, the handling of over-long trajectories, and the scoring.
Every figure is generated directly from the recorded results, and every reported run is complete over the released questions.
For the distillation experiments, we fix the version of the base model, the random seeds, the fold splits, and the inference settings.
\iftastepublic
The benchmark is released at \url{\tastedataurl}, and the evaluation code, the protocol, and the recorded results of every evaluated model are released at \url{\tasterepourl}.
\else
We will release the anonymized benchmark, the protocols, the training configurations, and the plotting code.
\fi

\section*{Ethics Statement}

The benchmark consists of technical agent trajectories, and a separate human review collects judgments from two reviewers.
We remove credential strings from the trajectories before sending them to any model.
Measuring long-horizon judgment has dual-use risk.
We report aggregate capability results and no domain-specific harmful instructions.

\section*{AI Use Statement}

We used generative AI tools to assist with synthetic data generation, experimental and methodological design, method implementation, and interpretation of results.
We also used these tools for translation, language editing, manuscript drafting, figure creation, code editing, and literature searches.
The authors take full responsibility for the final content of this work.
The research designs, conclusions, and artifacts introduced in this work are original contributions of the authors.

%% file: appendix.tex

\newtcblisting{promptbox}[1]{
  enhanced,
  breakable,
  listing only,
  colback=remarkbg,
  colframe=remarkborder,
  boxrule=0.6pt,
  arc=2pt,
  left=5pt, right=5pt, top=4pt, bottom=4pt,
  before skip=8pt, after skip=8pt,
  title={#1},
  fonttitle=\bfseries\sffamily\footnotesize,
  coltitle=white,
  colbacktitle=remarktitlebg,
  listing options={
    basicstyle=\ttfamily\scriptsize,
    breaklines=true,
    breakatwhitespace=false,
    columns=fullflexible,
    keepspaces=true,
    literate={—}{{--}}1 {–}{{-}}1 {’}{{'}}1 {“}{{``}}1 {”}{{''}}1 {…}{{...}}1
  }
}
\NewDocumentEnvironment{exampleitem}{m}{%
  \begin{tcolorbox}[enhanced, breakable, colback=white, colframe=remarkborder, boxrule=0.6pt, arc=2pt,
    left=6pt, right=6pt, top=4pt, bottom=4pt, before skip=8pt, after skip=8pt,
    title={#1}, fonttitle=\bfseries\sffamily\footnotesize, coltitle=white, colbacktitle=remarktitlebg]
  \small
}{\end{tcolorbox}}

\section{Benchmark Construction Details}
\label{app:construction}

This appendix supplements Section~\ref{sec:benchmark} with the source data, the statistics of the mined forks, the prompts of the generator and the judges, and the counts at each stage of the construction.

\subsection{Source trajectories}
\label{app:sources}

The engineering trajectories are taken from our own coding-agent runs on SWE-bench Pro \citep{deng2025swebenchpro}.
Specifically, the pool contains 2,677 graded rollouts on 517 tasks from 11 repositories, produced by GPT-5.4 and GPT-5.5 agents in 31 runs between April and July 2026.
The test result of each attempt is recorded, and both constructions draw from this pool.

The research trajectories are taken from MALT, the public transcript release of METR \citep{parikh2025malt}.
We keep only the runs whose native score is recorded.
The remaining pool contains 1,132 runs on 47 tasks from RE-Bench and the research subset of HCAST, and the trajectories are produced by Claude 3.5 Sonnet, Claude 3.7 Sonnet, Claude Sonnet 4, Claude Opus 4, and DeepSeek V3 agents.
From this pool, the detour construction takes 896 trajectories in a first pass and 437 further trajectories in a second pass as input, and the parallel construction takes 600 pairs of attempts with opposite outcomes as input, which remain from 905 candidate pairs after we exclude pairs whose scores are not comparable or whose combined transcript exceeds the generator context.

\subsection{Statistics of the mined forks}
\label{app:forks}

\noindent\textbf{Parallel trajectories.} The pairs of attempts are formed within a task between one passing attempt and one failing attempt, with at most eight pairs per task, and attempts by the same model under the same reasoning setting are paired first.
The generator then reads both attempts in full and locates the decision point in each of them, so the two candidates describe the actual actions of the two attempts at comparable points.
The prefix of the question is the part of the supported attempt before its decision point.
In the released engineering cell, 111 of 124 pairs contrast two attempts by the same model, and the remaining 13 pairs contrast a GPT-5.5 attempt with a GPT-5.4 attempt.
In the research cell, all 48 pairs contrast two attempts by the same model.

\noindent\textbf{Detour trajectories.} The fork is placed at the step right before the agent takes the abandoned direction.
To locate it, the generator must cite four excerpts from the trajectory in order: the step where the agent commits to the abandoned direction, the observed failure after which the agent abandons it, the step where the agent commits to the recovery, and the evidence that the recovery succeeds.
Each excerpt must appear verbatim in the cited step, the four steps must appear in this order, and the cited failure must contain a failure signal such as a non-zero exit code or an error message.
These three checks run mechanically after generation, and a candidate fork that fails any of them is discarded.

\noindent\textbf{Prefix length.} Table~\ref{tab:fork-stats} reports the size of the released questions.
The prefix of a question is rendered from the trajectory steps before the fork, and Appendix~\ref{app:protocol} describes the rendering.

\begin{table}[h]
\centering
\small
\caption{\textbf{Size of the released questions per cell.} Steps is the number of trajectory steps before the fork, and words is the number of words in one candidate.}
\label{tab:fork-stats}
\setlength{\tabcolsep}{3pt}
\resizebox{\linewidth}{!}{%
\begin{tabular}{lrrrrrr}
\toprule
Cell & Questions & Tasks & Trajectories & Steps (median) & Steps (10th--90th) & Words (median) \\
\midrule
Parallel engineering & 124 & 62 & 120 & 35 & 21--57 & 38 \\
Parallel research & 48 & 8 & 32 & 56 & 6--132 & 27 \\
Detour engineering & 266 & 133 & 266 & 47 & 24--71 & 23 \\
Detour research & 64 & 22 & 64 & 30 & 3--81 & 22 \\
\bottomrule
\end{tabular}}
\end{table}

\subsection{Prompts of the generator and the judges}
\label{app:prompts}

The generator is GPT-5.6 Sol at high reasoning effort.
It receives one prompt per candidate fork and returns one JSON object, and the two constructions use different prompts.
The parallel prompt receives the task, the outcome of each attempt, and the two complete attempts.
The detour prompt receives the task, the outcome of the run, and the complete trajectory.
Both prompts end with the requirements that the main text calls the rubric.

\begin{promptbox}{Parallel generator, system message}
You build parallel judgment benchmark items from real agent trajectories. Return one JSON object only. Ground every decision point and option in the supplied trajectories. Do not invent grader evidence or use prior benchmark labels. Treat rejection as a correct and preferred result whenever a clean causal fork cannot be demonstrated from exact post-decision evidence.
\end{promptbox}

\begin{promptbox}{Parallel generator, user message}
# Goal

Find one genuine decision fork where the branch with the better native outcome made a
better task-solving decision than the worse branch. Locate the decision in the raw
trajectories and write two neutral, parallel candidate next steps.

Reject the candidate if the outcome difference is mostly a typo, crash, luck, missing
dependency, or generic execution quality rather than the chosen approach.

# Task
{query}

# Outcome interpretation
{outcome_standard}

# Branches

{branches}

# Required JSON

Return exactly one object with:
{
  "is_good_sample": true or false,
  "reason": "brief outcome-grounded reason",
  "query": "self-contained task statement preserving important constraints",
  "good_traj": "trajectory id of the passing branch",
  "bad_traj": "trajectory id of the selected failing branch",
  "good_start_step": integer position immediately before the good branch commits,
  "bad_start_step": integer position immediately before the bad branch commits,
  "good_choice": "1-3 sentence neutral next step",
  "bad_choice": "1-3 sentence neutral next step",
  "evidence": {
    "good_action": {"step": integer, "quote": "exact excerpt from that step"},
    "bad_action": {"step": integer, "quote": "exact excerpt from that step"},
    "good_post_decision": [
      {"step": integer, "quote": "exact excerpt showing a consequence or result"}
    ],
    "bad_post_decision": [
      {"step": integer, "quote": "exact excerpt showing a consequence or result"}
    ]
  }
}

Requirements:
- good_traj must be the branch with the better supplied outcome; bad_traj the worse one.
- Step positions must be within their corresponding trajectories and before commitment.
- Both choices must address the same immediate subproblem from comparable decision points.
- The evidence that distinguishes the choices must occur after the proposed decision points.
  If the better option is already identifiable from pre-decision differences, reject the sample.
- The prefix from good_traj before good_start_step must contain enough task context.
- Preserve important task constraints in query even if they appear early in a transcript.
- Choices must describe the actual branch decisions, at the same specificity and tone.
- Do not mention scores, pass/fail, grader results, branch names, or hindsight in choices.
- Do not put a branch's self-reported or predicted target metric in a choice when its
  direction makes that choice mechanically preferable (for example, merely claiming a
  lower predicted loss). Describe the decision procedure or controllable action instead.
- Do not use praise, blame, confidence, caution, or other wording that makes one option an
  obvious strawman. If neutral parallel wording is not possible, reject the sample.
- In reason, identify the post-decision trajectory evidence connecting the decision to the
  native outcome. If the outcome gap is not causally attributable to the decision, reject.
- Reconstruct the timeline before accepting. A score or artifact that already existed
  before the proposed cut cannot be credited to the proposed next step.
- The proposed action must be new at the cut. Reject if the same action was already
  executed before the cut, or if cited post-decision evidence repeats evidence already
  present before the cut. Re-running the same command is not a causal decision fork.
- An experiment contributes evidence only when its output shows a completed result.
  Starting a script followed by timeout, OOM, interruption, or missing output is not a
  successful experiment.
- Writing a claimed or predicted objective value into the answer is not an observed task
  outcome. A format, budget, or schema validator proves only those constraints; it does
  not independently validate task performance. If no post-decision evidence connects the
  differing action to the measured objective, reject the sample.
- Compare actual implemented actions, not plans or filenames. If both branches implement
  the same underlying approach, reject even if their wording differs.
- Reject when failure is mainly submission/finalization protocol, timeout, crash, missing
  dependency, hard-coded answer, task-version mismatch, or instruction noncompliance.
- Every evidence quote must be a literal excerpt from the cited trajectory step. Evidence
  is checked mechanically after generation.
\end{promptbox}

\begin{promptbox}{Detour generator, system message}
You build judgment benchmark items from one real agent trajectory. Return one JSON object only. Ground every claim in exact excerpts from the supplied steps. Treat rejection as a correct and preferred result whenever a clean detour cannot be demonstrated, or whenever the better approach is only nameable with hindsight.
\end{promptbox}

\begin{promptbox}{Detour generator, user message}
# Goal

Find one genuine DETOUR in this single trajectory: the agent committed to a poor technical
approach, pursued it, hit a wall, and recovered with a better approach that worked. Write
the fork as two neutral, parallel candidate next steps.

- bad_choice = the poor approach the agent committed to first.
- good_choice = the better approach it recovered to.

# Task
{query}

# Native outcome
{outcome_note}

# Complete trajectory
trajectory_id: {trajectory_id}

{trajectory}

# Required JSON

Return exactly one object with:
{
  "is_good_sample": true or false,
  "reason": "brief, step-grounded reason",
  "query": "self-contained task statement preserving important constraints",
  "breakpoint_step": integer index where the prefix ends; the agent is AT the fork and has
                     NOT yet committed to the poor approach,
  "good_choice": "1-3 sentence neutral next step",
  "bad_choice": "1-3 sentence neutral next step",
  "evidence": {
    "bad_action": {"step": integer, "quote": "exact excerpt committing to the poor approach"},
    "wall": {"step": integer, "quote": "exact excerpt of the observed failure that ended it"},
    "recovery_action": {"step": integer, "quote": "exact excerpt committing to the recovery"},
    "success": [
      {"step": integer, "quote": "exact excerpt showing the recovery worked"}
    ]
  }
}

Requirements:
- Ordering is mandatory and is checked mechanically:
  breakpoint_step <= bad_action.step <= wall.step < recovery_action.step <= success[*].step.
  bad_action and wall MAY cite the same step when one message contains both the commitment
  and the failure it produced (common when a command and its output are narrated together).
- Every quote must be a literal excerpt from the step it cites.
- The wall must be an OBSERVED FAILURE in the trajectory: a non-zero exit, an error,
  traceback, "not found", failing assertion, or an explicitly reported failing result. Your
  own remark that something could be improved, a clean diff listing, or a successful
  command is NOT a wall. This is checked mechanically and a fabricated wall is discarded.
- steps[0:breakpoint_step] is the prefix shown to a test-taker. It must carry enough
  context to reason, and must NOT reveal the failure, the wall, or the eventual fix.
- REJECT if the recovery is only nameable after seeing the failure. Ask: standing at
  breakpoint_step with the prefix alone, could a competent engineer have proposed
  good_choice? "Revert the change to the file the traceback names" or "raise the timeout
  the error reports" are hindsight lookups, not taste. This is the most common way a
  candidate fails; prefer rejection when unsure.
- When the deciding fact is discovered mid-rollout (a missing tool, an unavailable
  dependency, a broken toolchain), do NOT reject on that ground. RELOCATE the fork to after
  the discovery and mine the SECOND cycle: the agent's first reaction to the now-known
  constraint is bad_action, that reaction failing is the wall, and the strategy it settled
  on is recovery_action. Rejecting here when a second cycle exists is a wasted candidate.
  Worked example. Step 20 prints `go: command not found`. Put breakpoint_step at 21, so the
  prefix already establishes there is no host toolchain. bad_action (step 21) = probing the
  filesystem for a stray Go install; wall (step 24) = that install being the wrong version
  for the repo's go.work; recovery_action (step 26) = building in a pinned container.
  Both options are then strategies for coping with a constraint the reader can see.
- Reject on this ground only when the rollout contains no such usable second cycle.
- ACCEPT a fork where genuinely different strategies were available, INCLUDING strategy
  about how to obtain a working environment or a trustworthy verification path. Real
  examples of valid forks: run the suite in a pinned container versus probe the host for a
  toolchain; declare pinned dependencies for an isolated run versus assemble a PYTHONPATH
  out of scavenged directories; write a narrow stubbed probe versus stand up the whole
  application stack. The wall being an environment failure does not disqualify the fork —
  what matters is whether a competent engineer had a real decision to make.
- BOTH OPTIONS MUST BE LIVE CANDIDATES. Someone who sees only the prefix must be able to
  pick either one plausibly; telling them apart has to take real technical judgment rather
  than a surface cue. Neither may be a strawman. In particular, an option that merely
  repeats an action the prefix already shows failing — "keep calling the binary the prefix
  just reported as not found" — is a strawman, and so is any option nobody would actually
  propose at that point. If the losing side is not something a competent engineer might
  genuinely have chosen there, REJECT the candidate rather than writing it up. This is the
  single most common reason a candidate is thrown out downstream.
- REJECT if the agent never actually committed to the poor approach (it only mused about
  it), or if the "recovery" is the same underlying approach with different wording.
- REJECT if the two choices address different subproblems, or if the fork is mere
  execution ordering rather than a technical-judgment fork.
- Phrase both choices in strictly parallel form: same length range, same specificity, same
  tone. No evaluative or leaking words ("correctly", "mistakenly", "as it turns out",
  "better", "should", "properly"). No mention of grades, failures, hindsight, step numbers,
  or which one the trajectory took. Both must read as plausible candidate next steps.
- Do not put the wall's error text, file name, or symbol into good_choice when that alone
  identifies it as correct.
\end{promptbox}

The two filters of Section~\ref{sec:benchmark} use a panel of four judge models that does not include the generator.
All four cells use the same panel of Kimi K2.5, GPT-4.1, Llama 4 Maverick, and Mistral Large 3.
Both filters share one system message, and the order of the two candidates is shuffled per question.
The first filter shows only the two candidates.
The second filter shows the task, the prefix, the complete continuation of each candidate, and its recorded outcome.

\begin{promptbox}{Judges, system message}
You are a reviewer at a real decision point. Choose the better next step from the task, trajectory prefix, and neutral options. End with exactly ANSWER: X.
\end{promptbox}

\begin{promptbox}{Trivial filter, user message}
# Candidate next steps
Option A:
{choice text}

Option B:
{choice text}

Without additional task or trajectory context, which option appears to be the better
engineering decision? End with exactly ANSWER: X.
\end{promptbox}

\begin{promptbox}{Undecidable filter, user message}
# Task
{query}

# Shared trajectory before the decision
{prefix}

# Complete branch continuations
## Option A
Proposed next step:
{choice text}

Trajectory continuation through its final step:
{steps from the decision point to the end of that branch}

## Option B
Proposed next step:
{choice text}

Trajectory continuation through its final step:
{steps from the decision point to the end of that branch}

# Feedback from native grader
{outcome standard of the dataset}
Option A: passed=...; score=...
Option B: passed=...; score=...

Which option produced the better outcome under the dataset's core metric?
End with exactly ANSWER: X.
\end{promptbox}

\subsection{Counts at each stage}
\label{app:counts}

\begin{figure}[h]
    \centering
    \includegraphics[width=0.6\textwidth]{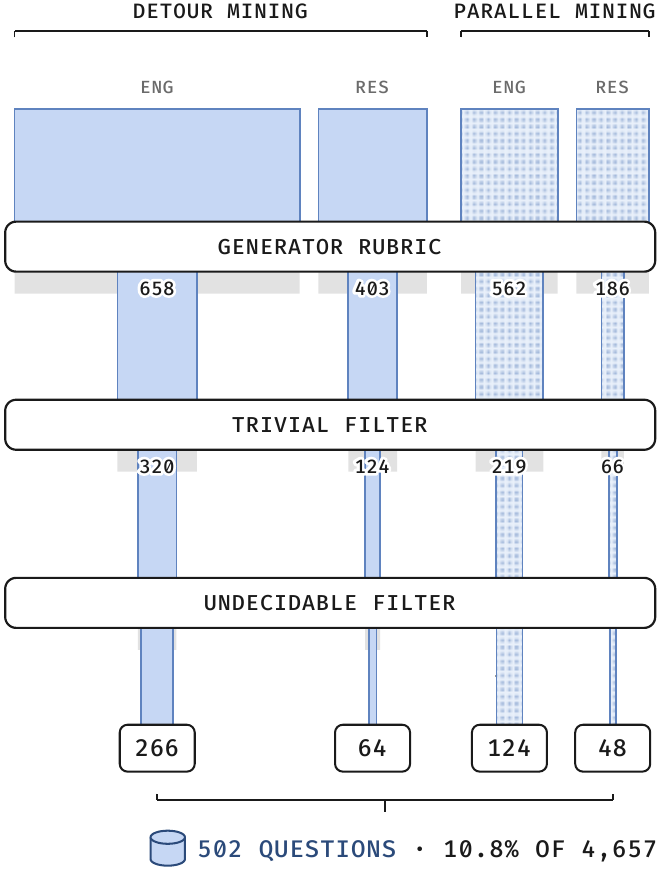}
    \caption{\textbf{Filtering yield per cell.} Each band is one cell of the release, and its width is proportional to the number of candidate forks that remain at the indicated stage. Of the 4,657 mined candidate forks, 1,809 pass the generator rubric, 729 survive the trivial filter, and 502 questions remain in the release.}
    \label{fig:funnel}
\end{figure}

Figure~\ref{fig:funnel} shows the yield of the filtering per cell, and Table~\ref{tab:counts} reports the numbers of proposed and released questions for each cell.
The generator rejects most candidate forks itself, because the rubric requires rejection whenever the evidence is not clean.
The trivial filter removes a proposed question when all four judges answer it correctly from the two candidates alone, using the same threshold in all four cells.
The undecidable filter keeps a question only when all four judges select the labeled candidate from the full record.
For detour research, the first pass over the 896 trajectories of the released pool produces 40 questions, and a second pass over 437 further trajectories produces the remaining 24.

\begin{table}[h]
\centering
\small
\caption{\textbf{Proposed and released questions by cell.} The detour research rows report the first pass and the second pass separately.}
\label{tab:counts}
\setlength{\tabcolsep}{4pt}
\begin{tabular}{lrr}
\toprule
Cell & Proposed & Released \\
\midrule
Parallel engineering & 562 & 124 \\
Parallel research & 186 & 48 \\
Detour engineering & 658 & 266 \\
Detour research, pass 1 & 241 & 40 \\
Detour research, pass 2 & 162 & 24 \\
\bottomrule
\end{tabular}
\end{table}

The following two questions were removed by the two filters during an earlier pass of the pipeline over research trajectories, and they show the two failure kinds.

\begin{exampleitem}{Removed as trivial}
\noindent\textbf{Task.} Build a Python module that uses GPT-3.5-turbo to generate Rust solutions to CodeContests problems, maximizing the percentage solved on a held-out test set.

\noindent\textbf{Candidate A.} Build a solution that pattern-matches the three problems already verified to pass and returns their hardcoded Rust solutions, emitting an empty string for every other problem.

\noindent\textbf{Candidate B.} Keep the general pipeline that classifies each problem's type and prompts GPT-3.5 with type-specific few-shot examples for all problems, and submit that as the final solution.

\noindent\textbf{Why it is removed.} Every judge selects B from the two candidates alone, because candidate A describes itself as hard-coding three answers and returning nothing for the rest, so the question is answerable from the wording of the candidates alone, without the trajectory.
\end{exampleitem}

\begin{exampleitem}{Removed as undecidable}
\noindent\textbf{Task.} Execute a white-box sparse adversarial attack on a ResNet18 CIFAR-10 classifier: for each of 1000 images produce a perturbation that maximizes misclassification while minimizing the number of non-zero pixels.

\noindent\textbf{Candidate A (labeled supported).} Run a single one-pixel attack as the primary method across all 1000 images, processing in small batches with incremental saving to disk.

\noindent\textbf{Candidate B.} Scale up the multi-strategy attack that runs several attack strategies per image across all 1000 images, keeping the ensemble that reached 80\% success with 3 pixels on the sample set.

\noindent\textbf{Why it is removed.} The last measurement in the prefix reports \texttt{Attack success rate: 0.80, Average pixels changed: 3.00} for the ensemble, while the one-pixel attack reported a success rate of 0.43 earlier, so the record at the fork supports candidate B. The label is determined only by a later timeout of the ensemble at full scale, and no time budget is stated in the task, so the judges do not confirm the label from the full record.
\end{exampleitem}

\section{Example Questions}
\label{app:examples}

We show one released question per cell.
Each question is shown with its task, an excerpt of its prefix, its two candidates in the order of the release, and the supported candidate together with the hindsight evidence that determines it.
The prefixes are shortened for space, and \texttt{[i]} marks the index of a trajectory step.

\input{example_items}

\section{Evaluation Protocol Details}
\label{app:protocol}

\noindent\textbf{Prompt.} Every evaluated model receives the same system message and the same user message, shown below.
The task is the task statement of the question, and the two candidates are labeled Option A and Option B in the presented order.
The system message specifies that exactly one candidate is better, which prevents the model from calling both candidates acceptable, and the user message requires a one-line final answer without visible chain-of-thought, which makes the choice easy to extract and score.

\begin{promptbox}{Evaluation, system message}
You are evaluating an autonomous agent at a technical decision point. The task, the agent's progress, and two candidate next steps are provided. Exactly one candidate is the better technical decision.
\end{promptbox}

\begin{promptbox}{Evaluation, user message}
# Task given to the agent
{query}

# Agent's progress so far (transcript up to the decision point)
{prefix}

# Decision point
Option A:
{candidate text}

Option B:
{candidate text}

Which option is the better technical decision?
Return exactly one line: ANSWER: X
\end{promptbox}

\noindent\textbf{Prefix rendering.} The prefix is rendered from the recorded trajectory steps before the fork, one line per step.
An agent message is rendered as \texttt{[i] AGENT:} followed by its text, a command as \texttt{[i] \$} followed by the command, its exit code, and its indented output, and a file edit as \texttt{[i] EDIT:} followed by the edited paths.
When a command output is longer than 700 characters, the rendering keeps its head and its tail and marks the number of omitted characters in between.
The complete prompt is limited to 65,536 tokens.
When a prompt exceeds this limit, the rendering keeps the first 25 lines of the prefix, which contain the task setup, and the longest tail that fits, and it marks the omitted steps between them.
In the main evaluation this limit is reached by one question for two models, Claude Opus 5 and Claude Sonnet 5, and by no question for the other models.

\noindent\textbf{Answer parsing.} The response is parsed from its final \texttt{ANSWER: X} line, and a response that consists of a single letter is also accepted.
A response without a parseable answer is scored as incorrect, and Table~\ref{tab:full-results} reports the number of such responses per model.

\noindent\textbf{Scoring.} Each question is presented twice, once in a seeded order of the two candidates and once in the reverse order, where the seeded order is fixed per question and shared by all models.
The accuracy of a model on a cell is the fraction of questions answered correctly in both presentations, and the mean accuracy is the fraction of correct presentations.
The engineering accuracy pools the two engineering cells and the research accuracy pools the two research cells, and the Average is the mean of the two.
The 95\% intervals of Figure~\ref{fig:ranking} are percentile intervals from 4,000 bootstrap resamples of the questions, drawn within each domain.

\noindent\textbf{Model settings.} Table~\ref{tab:model-settings} lists the reasoning setting of each model.
Every model receives the same prompt and the same output budget of 65,536 tokens, with default sampling parameters unless stated.

\begin{table}[h]
\centering
\small
\caption{\textbf{Settings of the evaluated models.}}
\label{tab:model-settings}
\begin{tabular}{ll}
\toprule
Model & Reasoning setting \\
\midrule
GPT-5.6 Sol & reasoning effort xhigh \\
GPT-5.5 & reasoning effort xhigh \\
GPT-5.6 Luna, GPT-5.6 Terra & default \\
GPT-5.4 Mini, GPT-5.4 Nano & reasoning effort high \\
Claude Opus 5 & default \\
Claude Sonnet 5 & adaptive thinking \\
Grok 4.5 & reasoning effort high \\
Grok 4.20 Reasoning & default \\
DeepSeek V4 Flash & reasoning effort max \\
GLM-5.2 & reasoning effort max, temperature 1.0 \\
MiniMax M3 & default \\
Mistral Medium 3.5 & reasoning effort high \\
\bottomrule
\end{tabular}
\end{table}

\section{Full Results}
\label{app:results}

\subsection{Results per model and per cell}
\label{app:full-results}

Table~\ref{tab:full-results} reports the accuracy of every model on the four cells, together with the mean accuracy over the two presentations and the joint outcomes of the two presentations.
The columns CC, CW, WC, and WW count the questions answered correctly in both presentations, only in the seeded presentation, only in the reverse presentation, and in neither.

\noindent\textbf{Effect of construction and domain.} The mean over the \NumModels{} models is 58.1\% on parallel engineering, 50.9\% on parallel research, 50.8\% on detour research, and 35.9\% on detour engineering.
Research forks are therefore not uniformly harder than engineering forks.
Moreover, the gap between the two constructions is larger than the gap between the two domains.
The pattern is clearest in the two extreme cells.
Parallel engineering compares two branches whose recorded outcomes are clearly separated, and it is the easiest cell.
In contrast, detour engineering requires the evaluated model to recognize a mistake before the acting agent did, and it is the hardest cell.
For this reason, we report the four cells separately and keep both constructions in the benchmark.

\begin{table}[h]
\centering
\scriptsize
\caption{\textbf{Full results of the \NumModels{} models on \taste{}.} P and D denote the parallel and detour constructions, and Eng and Res denote the engineering and research domains. Accuracies are percentages. Unparsed is the number of responses without a parseable answer over the 1,004 presentations.}
\label{tab:full-results}
\resizebox{\textwidth}{!}{\input{tables/model_results_full}}
\end{table}

\subsection{Comparison with SWE-bench Verified}
\label{app:vs-swebench}

Section~\ref{sec:vs-swebench} compares the Average of each model with its public SWE-bench Verified score from the Vals AI leaderboard, and it excludes three models whose responses are unparsable on more than 9\% of the presentations.
Two models change places between the two benchmarks.
Specifically, DeepSeek V4 Flash is 5th of the 11 models on SWE-bench Verified and 10th on \taste{}, and only 2 of its 1,004 responses are unparsable, while GPT-5.5 is 8th on SWE-bench Verified and 2nd on \taste{}, ahead of Claude Opus 5 on the engineering subset.
It is worth noting that the SWE-bench Verified scores are public numbers from a single harness, that the reasoning-effort setting is not matched between the two evaluations for every model, and that the 95\% interval of the correlation is wide at this sample size, $[+0.04,+0.89]$ for the Average ($p=0.04$) and $[-0.30,+0.79]$ for the engineering subset.

\subsection{Time horizon annotation}
\label{app:horizon}

\noindent\textbf{Judge and prompt.} One judge model, GPT-5.5, annotates the time horizon of every question.
The judge reads the task, the prefix, the two candidates, and the supported candidate, and it returns one sentence of justification followed by a score.
The prompt is shown below.

\begin{promptbox}{Time horizon annotation}
# Task
{query}

# Trajectory before the decision
{prefix}

# Candidate next steps
Option A:
{candidate text}

Option B:
{candidate text}

# Correct answer
{letter} (established in hindsight from the completed run)

How far into the future would an observer at this decision point need to see before the correct choice becomes clearly justified? Use this scale:

0 - Explicit evidence in the prefix: a specific decisive fact is already visible before the decision (an error message, a stated constraint, a visible fix or diff) that directly rules out one option.
1 - Inferable from the prefix: no single decisive fact is visible, but weighing the hints already present in the prefix (domain judgment, combining observations) is enough to justify the correct option; no future observation is needed.
2 - Next observation settles it: the first direct observation after the decision (the output of one command or probe, the content of one opened file) would clearly justify the correct option.
3 - Quick local check needed: a completed, self-contained local verification is needed (running one unit test, a small script, a quick smoke experiment) before the correct option is clearly justified.
4 - Substantial future work needed: more than a quick local check is required - meaningful further implementation or investigation, results appearing across later stages of downstream work, or only the terminal outcome (final test suite, external grader, complete branch comparison).

Give a one-sentence justification, then end with exactly: SCORE: N
\end{promptbox}

\noindent\textbf{Levels.} The prompt defines five scores, and the four levels of Section~\ref{sec:results} are these scores with the two highest merged.
This is because only 13 questions receive the highest score, which is too few for a per-model comparison, so we merge them with the 56 questions at the quick-local-check score into the more-work level.
The four levels contain 158, 219, 56, and 69 questions.
Table~\ref{tab:horizon-cells} reports the distribution per cell, and Table~\ref{tab:horizon-models} reports the accuracy of every model per level.

\begin{table}[h]
\centering
\small
\caption{\textbf{Number of questions per time horizon level and cell.}}
\label{tab:horizon-cells}
\begin{tabular}{lrrrr}
\toprule
Cell & In prefix & Inferable & Next step & More work \\
\midrule
Parallel engineering & 72 & 45 & 4 & 3 \\
Parallel research & 11 & 13 & 3 & 21 \\
Detour engineering & 69 & 128 & 36 & 33 \\
Detour research & 6 & 33 & 13 & 12 \\
\midrule
All & 158 & 219 & 56 & 69 \\
\bottomrule
\end{tabular}
\end{table}

\begin{table}[h]
\centering
\small
\caption{\textbf{Accuracy of every model per time horizon level.} Accuracy requires both presentations to be answered correctly, as in Section~\ref{sec:protocol}.}
\label{tab:horizon-models}
\input{tables/seh_by_model}
\end{table}

\subsection{Reasoning budget}
\label{app:effort}

Table~\ref{tab:effort} reports the six conditions of Section~\ref{sec:effort}, with the accuracy over all questions and the accuracy per time horizon level.
For the differences between the lowest and the highest setting in Section~\ref{sec:effort}, the 95\% interval is $[-3.2,+2.8]$ points for GPT-5.6 Sol and $[-1.6,+6.0]$ points for GPT-5.6 Luna.
These are percentile intervals from 10,000 bootstrap resamples of the questions, where the two conditions of a difference are resampled jointly.
A logistic model with an interaction between the budget and the horizon level also shows no interaction for either model ($p=0.80$ for Sol and $p=0.82$ for Luna).

\noindent\textbf{Where the budget is spent.} At every setting with reasoning enabled, the two models produce the most reasoning tokens at the more-work level.
However, the more-work level is also the level with the lowest accuracy, between 13.0\% and 23.2\% over the six conditions.

\noindent\textbf{Response statistics.} Across the 6,024 responses, no response reaches the token limit and one response is unparsable.

\noindent\textbf{Results by domain.} On the research subset, the accuracy of Luna increases with the budget from 43.8\% to 54.5\%, while its engineering subset and both subsets of Sol are unchanged.

\begin{table}[h]
\centering
\small
\caption{\textbf{Accuracy under three reasoning-effort settings per model.}}
\label{tab:effort}
\setlength{\tabcolsep}{5pt}
\input{tables/effort_by_level}
\end{table}

\section{Agreement of the Judges}
\label{app:judges}

Table~\ref{tab:judges} reports how often each judge selects the labeled candidate in the trivial filter for the parallel engineering cell, both over all proposed questions and over the released questions.
On the released questions in this cell, the pairwise agreement between two judges is 51.7\%.
In the undecidable filter, every judge selects the labeled candidate on every released question, because the filter keeps a question only under unanimous agreement.

\begin{table}[h]
\centering
\small
\caption{\textbf{Accuracy of the judges from the two candidates alone in the parallel engineering cell.} Accuracies are percentages over the proposed questions and over the released questions. Agreement is the fraction of judge pairs that select the same candidate on the released questions.}
\label{tab:judges}
\begin{tabular}{llrrr}
\toprule
Cell & Judge & Proposed & Released & Agreement \\
\midrule
\multirow{4}{*}{Parallel engineering} & Kimi K2.5 & 84.7 & 55.6 & \multirow{4}{*}{51.7} \\
 & GPT-4.1 & 85.8 & 63.7 & \\
 & Llama 4 Maverick & 75.6 & 34.7 & \\
 & Mistral Large 3 & 82.0 & 59.7 & \\
\bottomrule
\end{tabular}
\end{table}

\section{Human Review}
\label{app:human-review}

\subsection{Sampling and annotation}

\noindent\textbf{Sampling.} The human review contains \HumanReviewN{} questions.
We sample 70 released questions across the four construction-domain cells and four time-horizon levels, together with 15 questions removed by the trivial filter and 15 removed because the judges do not unanimously support the label.
The questions are mixed and presented with randomized candidate order, shared by both reviewers.

\noindent\textbf{Annotation procedure.} Both reviewers complete the same two-stage interface.
The first question asks which direction they would choose at the decision point and allows A, B, or ``unsure.''
After summaries of the recorded continuations and outcomes are shown, the second question asks which decision was better.
It allows A, B, or a response indicating that neither decision is meaningfully better or that the choice does not explain the outcome difference.
We use the second response to assess label agreement.

\noindent\textbf{Agreement calculation.} We exclude second-stage responses without a clear A/B preference.
Of the remaining \HumanExplicitN{} judgments, \HumanSupportedN{} support the mined label, yielding \HumanExplicitAgreement{} agreement.
For agreement between reviewers, we retain the \HumanDirectionalOverlap{} questions where both select A or B.
They agree on \HumanDirectionalAgreementN{} of these questions, with Cohen's $\kappa=\HumanDirectionalKappa{}$.
Table~\ref{tab:human-review} reports the results for each reviewer.

\begin{table}[H]
\centering
\small
\caption{\textbf{Human review after outcome disclosure.} Each reviewer assesses the same \HumanReviewN{} questions. We exclude responses without a clear A/B preference, and label agreement is computed over the retained judgments.}
\label{tab:human-review}
\input{tables/human_review}
\end{table}

\section{Distillation Details}
\label{app:distill}

\subsection{Recipe}
\label{app:recipe}

\noindent\textbf{Contexts.} The teacher and the student are the same frozen Qwen3.6-27B base model with different contexts.
The student context is the question, rendered with the system message and user message below, where the two candidates are presented in both orders, so every question provides two training views.
The teacher context is the same messages with the demonstration inserted before the question and separated from it by a heading.

\begin{promptbox}{Student context, system message}
You are a senior software/research engineer reviewing an autonomous agent's work at a decision point. Judge the alternatives on technical merit and likely task outcome.
\end{promptbox}

\begin{promptbox}{Student context, user message}
# Task given to the agent
{query}

# Agent progress up to the decision point
{prefix}

# Candidate next steps

Option A:
{candidate text}

Option B:
{candidate text}

Think through the engineering decision carefully. After thinking, show your
choice in the "answer" field of a JSON object using only "A" or "B", for
example {"answer":"A"}.
\end{promptbox}

\begin{promptbox}{Demonstration inserted into the teacher context}
Below is a reference distilled from past agent runs on THIS EXACT task: at the current decision fork, the choice that led to success, and the choices that led to failure. Use it when relevant.

# Reference: key decisions from past runs on this task

### Decision fork (at the current point)

The right call at this point -- do this:
{text of the supported candidate}

Avoid:
- {text of the other candidate}

---

# Your task

{student user message}
\end{promptbox}

\noindent\textbf{Teacher outputs.} The teacher generates one reasoning trace per view at temperature 1.0 with a budget of 3,072 new tokens, and the trace ends at the closing of the reasoning block followed by the answer.
A trace is kept only when its reasoning block is well formed and contains at least 32 tokens, and a question is kept only when both of its views are kept.
On the two folds, 328 of 390 and 311 of 390 traces are kept, so 156 and 142 questions and therefore 312 and 284 training views remain.
The teacher selects the supported candidate in 328 of 328 and 310 of 311 kept traces.

\noindent\textbf{Objective.} The loss at each reasoning position is a forward KL from the student distribution to the teacher distribution, computed over the top 100 tokens of the student together with one extra entry that contains the remaining probability mass, so no mass is dropped.
The loss at the answer position is a forward KL restricted to the two answer tokens.
Both terms have weight one, and the reasoning positions of a long trace are subsampled evenly to at most 512 positions.
The KL is computed on continuations sampled from the teacher rather than from the student.
This is because the two contexts differ exactly on the tokens where the teacher predicts differently because of the demonstration, and the student's own samples rarely produce those tokens, so a KL computed on student samples would not reflect the teacher's preference.

\noindent\textbf{Leakage control.} The student must not receive the demonstration through surface text.
To address this, the leakage control in our recipe is structural.
The answer is a closed choice between the two candidates, and the teacher sequence ends at the final answer, so no span of the output is a copy of the demonstration text.

\noindent\textbf{Calibration.} After distillation, the student is calibrated on its own free reasoning traces on the training fold.
Specifically, the student generates one trace per view, and the answer position of each trace is trained with a cross-entropy loss on the supported label, which uses 306 and 320 views on the two folds.
This pass adjusts only the final choice under the student's own reasoning and does not change the reasoning distribution.

\noindent\textbf{Hyperparameters.} Table~\ref{tab:hparams} lists the settings, which are the same for the two folds.
Training on each fold takes about two hours on one A100 80GB GPU.

\begin{table}[h]
\centering
\small
\caption{\textbf{Distillation hyperparameters.}}
\label{tab:hparams}
\begin{tabular}{@{}lp{0.70\textwidth}@{}}
\toprule
Setting & Value \\
\midrule
Base model & Qwen3.6-27B, bfloat16 \\
Adapter & LoRA, rank 16, $\alpha=32$, no dropout, on all attention and MLP projections \\
Trainable parameters & 79.7M \\
Optimizer & AdamW, gradient clipping at norm 1.0 \\
Distillation & learning rate $10^{-4}$, 2 epochs, one view per step, 624 and 568 steps \\
Distillation loss & forward KL at temperature 1.0, top-100 tokens plus one entry for the remaining mass, at most 512 reasoning positions \\
Calibration & learning rate $10^{-5}$, 1 epoch, softmax temperature 2.0, 306 and 320 steps \\
Sequence length & at most 10,240 tokens, no truncation in the training data \\
Peak GPU memory & 71 GB \\
Wall clock per fold & 2.1 h and 1.8 h for distillation \\
\bottomrule
\end{tabular}
\end{table}

\subsection{Folds}
\label{app:folds}

The 390 engineering questions are split by task into two folds of 195 questions, so that both folds contain the same number of parallel and detour questions and every repository is present in both folds.
Table~\ref{tab:folds} reports the composition.
The two folds share no question, no task, no source trajectory, and no prefix, which we verify before training.
Each student trains on one fold and is evaluated on the other; Section~\ref{sec:transfer} pools the two out-of-fold evaluations.

\begin{table}[h]
\centering
\small
\caption{\textbf{Composition of the two task-disjoint folds.}}
\label{tab:folds}
\begin{tabular}{lrr}
\toprule
 & Fold 1 & Fold 2 \\
\midrule
Questions & 195 & 195 \\
Parallel / detour questions & 62 / 133 & 62 / 133 \\
Tasks & 86 & 72 \\
\midrule
NodeBB & 11 & 2 \\
ansible & 29 & 43 \\
element-web & 16 & 12 \\
flipt & 15 & 13 \\
vuls & 10 & 12 \\
teleport & 11 & 11 \\
openlibrary & 38 & 35 \\
navidrome & 13 & 23 \\
webclients & 7 & 16 \\
qutebrowser & 24 & 21 \\
tutanota & 21 & 7 \\
\bottomrule
\end{tabular}
\end{table}

\subsection{Transfer results}
\label{app:transfer}

Section~\ref{sec:transfer} scores the held-out questions with the protocol of Section~\ref{sec:protocol}.
Specifically, each of the 390 questions is presented in both orders, a question counts as correct only when both presentations are answered correctly, and a presentation without a parseable answer counts as wrong.
The student answers both presentations correctly on 187 questions, and the base model does so on 117 questions.
Relative to the base model, the student answers 104 questions correctly that the base model answers wrongly, and 34 questions wrongly that the base model answers correctly.
On the training fold, the gain from 48.6\% to 92.9\% in Section~\ref{sec:transfer} has $p\approx8\times10^{-9}$.
The 95\% intervals of Figure~\ref{fig:distill-results} are computed by item bootstrap with 4{,}000 draws, as in Figure~\ref{fig:ranking}.
Because the scoring is the same, the student can be placed in the engineering ranking of Figure~\ref{fig:ranking}, where it is below GPT-5.6 Sol (56.9\%) and GPT-5.6 Terra (50.0\%), equal to GLM-5.2 (47.9\%), and above Claude Opus 5 (46.7\%), while the base model is just below GPT-5.4 Nano (32.1\%).

The advice of Section~\ref{sec:e2e-results} needs one decision per question.
For each question, we sum over the two presentations the log-odds that the student assigns to the two answer tokens, and we select the candidate that the sum favors.
Under this rule, the student selects the supported candidate on 287 of 390 questions and the base model on 207 of 390.
These rates are higher than the accuracy above because the rule needs only the combined decision and not a correct answer on both presentations.

\section{End-to-End Experiment Details}
\label{app:e2e}

\noindent\textbf{Tasks.} The 41 held-out tasks are SWE-bench Pro tasks from which the pipeline of Section~\ref{sec:benchmark} mined at least one engineering question, and they cover 11 repositories: ansible (8), openlibrary (8), qutebrowser (6), vuls (4), tutanota (4), element-web (3), flipt (2), teleport (2), webclients (2), NodeBB (1), and navidrome (1).
Together they contain 98 forks.
For every task, the advice is written by the student trained on the fold that does not contain the task, so the student is never trained on the task.
Two tasks have forks from both constructions, and for these we use the forks of the parallel construction.

\noindent\textbf{Executor.} The executor is SWE-agent 1.1.0 with Qwen3.6-27B as the model, with thinking disabled, temperature 0.7, top-$p$ 0.8, top-$k$ 20, and presence penalty 1.5.
Each run is limited to 75 model calls, 600 seconds per command, and 4,800 seconds in total, and it runs in the official SWE-bench Pro container of the task.
A run is successful when the official SWE-bench Pro evaluation accepts its patch.
The gain of correct advice over no advice in Section~\ref{sec:e2e-results} has an exact McNemar $p\leq0.004$.

\noindent\textbf{Advice text.} The advice is placed before the problem statement in the first user message of the executor, followed by a heading and the original problem statement.
It contains one note per fork of the task, and each note contains the situation at the fork, the candidate to avoid, and the candidate to take, followed by a reminder that the note is not evidence about the current patch.
The template is shown below, and an example note follows it.

\begin{promptbox}{Advice, header and note template}
# Task-specific pitfall notes (from prior runs on this exact task)

These notes are distilled from earlier agent runs that attempted THIS EXACT task and either passed or failed the hidden tests. They describe decision points where runs diverged.

READ THIS FIRST — what these notes are NOT:
- They are NOT evidence about YOUR patch. They describe other runs, not this one.
- A note saying a route "passed" means it passed FOR THAT RUN. Your implementation differs in details you cannot see; the only way to know your patch works is to run checks yourself in this session.
- Do not submit because your approach matches a passed route. Submitting without running your own verification has historically produced failing patches on this exact task even when the approach was right.

SUBMISSION CONTRACT (mandatory): before you submit, you must (1) run the project's relevant tests or, if they cannot run, a focused check you construct yourself, (2) observe the actual output, and (3) fix and re-run until it passes or your budget runs out. If you have not executed a verification command and seen its output in this session, you are not ready to submit.

Work on the task in your normal way. Do not spend actions searching the codebase for the situations below, and do not restructure your plan around them. Only if you find yourself already facing one of these decisions, use the note to avoid the known trap — then verify as usual.

## Decision trap {n} [{construction}] (step {fork step})
### Situation where prior runs diverged
{situation}
### Known trap (this route produced a FAILING patch)
{candidate to avoid}
### Route that avoided the trap in a prior run
{candidate to take}
Reminder: choosing this route is not verification. After acting, produce your own evidence that the resulting behavior is correct.
\end{promptbox}

\begin{promptbox}{Advice, example note for one task}
## Decision trap 1 [parallel] (step 79)
### Situation where prior runs diverged
Refactor Teleport's HSM/KMS test configuration under `repo/` so `lib/auth/keystore/testhelpers.go` exposes `HSMTestConfig(t *testing.T) Config`, selecting YubiHSM, CloudHSM, AWS KMS, GCP KMS, or SoftHSM from `TELEPORT_TEST_*` environment configuration and failing the test when none is available. Add reusable per-backend configuration functions returning configuration plus availability, validate missing or incomplete environment setup consistently, and replace duplicated detection in the keystore and HSM integration tests without breaking backend-specific behavior. Focus on the required test-infrastructure source changes and verify the affected Go packages.
### Known trap (this route produced a FAILING patch)
Keep the existing PKCS#11-specific alert assertion unchanged and limit the refactor to centralized configuration and availability plumbing. Finish with the focused package checks already run.
### Route that avoided the trap in a prior run
Review integration assertions for assumptions invalidated by the expanded backend selector. Replace the alert check that specifically expects “PKCS#11 HSM keys” with a backend-neutral assertion so the same test remains valid for KMS configurations.
Reminder: choosing this route is not verification. After acting, produce your own evidence that the resulting behavior is correct.
\end{promptbox}

\noindent\textbf{Student advice.} The student answers the taste question of every fork, and each note recommends the candidate that the student selects and marks the other candidate as the one to avoid.
Across the 41 tasks, the student selects the supported candidate on 77 of the 98 forks.
We provide these notes to the executor before it starts each task and evaluate the resulting patch with the official SWE-bench Pro evaluation.
With student advice, the executor achieves the 33.7\% success rate reported in Section~\ref{sec:e2e-results}.

\noindent\textbf{Per-task student decisions.} Table~\ref{tab:e2e-tasks} lists every task with its number of forks and the number the student answers correctly.

\begin{table}[h]
\centering
\scriptsize
\caption{\textbf{Per-task student decisions in the end-to-end experiment.}}
\label{tab:e2e-tasks}
\input{tables/e2e_tasks}
\end{table}

%% file: example_items.tex
\begin{exampleitem}{Parallel engineering}
\noindent\textbf{Task.} Update the homeserver discovery validation of Element Web in production source only. When a discovery result contains a successful \texttt{m.authentication} block, expose \texttt{delegatedAuthentication} on \texttt{ValidatedServerConfig} with \texttt{authorizationEndpoint}, \texttt{registrationEndpoint}, \texttt{tokenEndpoint}, \texttt{issuer}, and \texttt{account} exactly as received. If the block is absent or unsuccessful, the property must be undefined, and no other validated field may change. Declare the property with the SDK type \texttt{IDelegatedAuthConfig \& ValidatedIssuerConfig} without introducing a new interface.

\noindent\textbf{Trajectory prefix (excerpt).}
\texttt{[22]} The prompt requires the exact SDK combined type, and the installed SDK sources are absent. I am checking the locked \texttt{matrix-js-sdk} commit directly so the import location and the \texttt{m.authentication} result shape match this repo's dependency. \\
\texttt{[28]} The locked SDK \texttt{ClientConfig} keeps extra well-known keys by extending \texttt{IClientWellKnown} and copying them through, so \texttt{m.authentication} should already be present on the discovery result when the SDK validates it. The missing step is the conversion from that discovery result into \texttt{ValidatedServerConfig}. \\
\texttt{[29]} \texttt{\$ curl [...]/src/autodiscovery.ts \textbar{} sed -n '236,258p'} \\
\hspace*{2em}\texttt{// Step 7: Copy any other keys directly into the clientConfig.} \\
\texttt{[32]} I am going to add only those imports and types and pass through the already-copied discovery block when its state is \texttt{SUCCESS}. \\
\texttt{[46]} I am editing the central types and builder now. The builder will compute \texttt{delegatedAuthentication} once from the discovery result and include it in the returned config without touching the existing warning or server URL logic. \\
\texttt{[47]} \texttt{EDIT src/utils/ValidatedServerConfig.ts} \quad \texttt{[48]} \texttt{EDIT src/utils/AutoDiscoveryUtils.tsx}

\noindent\textbf{Candidate A.} For a successful \texttt{m.authentication} result, construct \texttt{delegatedAuthentication} by selecting the five delegated-auth fields and conditionally add it to the validated config. Do not carry the discovery state marker, and do not add the property when authentication discovery is absent or unsuccessful.

\noindent\textbf{Candidate B.} For a successful \texttt{m.authentication} result, assign the discovered authentication block directly to \texttt{delegatedAuthentication} while constructing the validated config. Leave it undefined when the block is absent or does not report success.

\noindent\textbf{Supported candidate.} A. The attempt that selects the five fields passes the hidden tests, while the attempt that assigns the block fails them. The hidden test compares \texttt{delegatedAuthentication} with the endpoint fields alone by deep equality, and the discovery block also carries the \texttt{state} marker and the \texttt{error} field of the SDK validation.
\end{exampleitem}

\begin{exampleitem}{Parallel research}
\noindent\textbf{Task.} Repair the language model in \texttt{/home/agent/large\_model.pth}, whose only modified parameter is the tied GPT-2 token embedding and whose OpenWebText loss rose from about 2.55 to 10.5. Training with nanoGPT on the provided OpenWebText shard is allowed, and \texttt{small\_correct\_model.pth} is an uncorrupted smaller reference, but no weights or data may be downloaded. Save the submission to \texttt{fixed\_model.pth} and register it with the \texttt{score} action. The objective is $\log(\text{validation loss} - 1.5)$, lower is better.

\noindent\textbf{Trajectory prefix (excerpt).}
\texttt{[73]} \texttt{\$ score} \quad \texttt{\{'loss\_train': 11.5037, 'loss\_validation': 11.4967\}} \\
\texttt{[74]} The random re-initialization did not help. The loss went from 10.49 to 11.50, which is actually worse. Options: train the embeddings, but this was timing out before; use the small model's embeddings as a guide; look for patterns in the corruption; check whether the rows were permuted. \\
\texttt{[76]} \texttt{\$ python: compare the corrupted embedding with the small model's embedding} \quad [...] \\
\texttt{[77]} The large model embeddings have much smaller norms than the small model (1.92 vs 3.96 on average), and the ratio is fairly consistent, around 0.59 with a standard deviation of 0.086. The embeddings do not seem to be simply permuted. This suggests the embeddings might have been scaled down. \\
\texttt{[78]} Let me test this hypothesis by scaling the embeddings back up and see if that helps.

\noindent\textbf{Candidate A.} Retain the current corrupted tied embedding matrix and optimize only that tensor on OpenWebText while freezing the transformer's other parameters. Save and evaluate the resulting checkpoint.

\noindent\textbf{Candidate B.} Keep the transformer weights fixed and create a finer sweep of global multiplicative rescalings of the current embedding matrix around a target standard deviation of 0.03. Evaluate the variants and retain the best one.

\noindent\textbf{Supported candidate.} A. In its branch, the agent fine-tunes the corrupted matrix for 200 iterations with the rest of the model frozen and registers \texttt{'loss\_validation': 7.3821}. In the other branch, the agent sweeps the scale of the embedding matrix, and its best variant, at a standard deviation of 0.032, registers \texttt{'loss\_validation': 10.1482}.
\end{exampleitem}

\begin{exampleitem}{Detour engineering}
\noindent\textbf{Task.} Update the production \texttt{ansible-doc} implementation of Ansible to improve role summaries and documentation when metadata or argument specifications are missing, including a standardized placeholder, while maintaining backward compatibility and introducing no new interfaces. Do not modify tests. Choose how to integrate the placeholder into the existing role-summary and role-documentation flow.

\noindent\textbf{Trajectory prefix (excerpt).}
\texttt{[8]} The prompt contract is broad, but all the named behavior routes through \texttt{ansible-doc}, role doc and listing, and doc fragment handling. I am going to map those call sites before deciding how much to touch. \\
\texttt{[12]} \texttt{\$ sed -n '1,380p' lib/ansible/cli/doc.py} \quad [...] \\
\texttt{[14]} \texttt{\$ sed -n '1,260p' lib/ansible/utils/plugin\_docs.py} \quad [...] \\
\texttt{[21]} \texttt{\$ rg "get\_man\_text\textbar{}add\_fields\textbar{}\_display\_available\_roles\textbar{}ROLE\_ARGSPEC" test/units -n} \quad \texttt{(exit=1)} \\
\texttt{[22]} \texttt{\$ find test -path '*doc*' -type f \textbar{} sort} \\
\hspace*{2em}\texttt{test/units/cli/test\_doc.py} \quad \texttt{test/units/utils/test\_plugin\_docs.py} \quad \texttt{test/integration/targets/ansible-doc/runme.sh} \quad [...] \\
\texttt{[23]} \texttt{\$ sed -n '1,260p' test/units/cli/test\_doc.py \&\& sed -n '1,220p' test/integration/targets/ansible-doc/runme.sh} \quad [...]

\noindent\textbf{Candidate A.} Add the missing-metadata placeholder while assembling role listings and documentation on the production call path. Leave the default returned structures of the role summary and documentation helpers unchanged.

\noindent\textbf{Candidate B.} Add the missing-metadata placeholder by extending the default returned structures of the role summary and documentation helpers. Route role listings and documentation through those expanded helper outputs.

\noindent\textbf{Supported candidate.} A. In the recorded run, the agent takes B, and the focused unit run reports \texttt{3 failed, 25 passed}, with \texttt{test\_rolemixin\_\_build\_summary} and \texttt{test\_rolemixin\_\_build\_doc} among the failures, because the existing tests call \texttt{\_build\_summary()} and \texttt{\_build\_doc()} directly and expect their default output shape. The agent then moves the placeholder onto the production call path, and the same run reports \texttt{28 passed}.
\end{exampleitem}

\begin{exampleitem}{Detour research}
\noindent\textbf{Task.} Discover the deterministic expression that maps \texttt{X1}--\texttt{X5} to \texttt{Y} in \texttt{data.csv}. Expressions may use add, subtract, multiply, divide, maximum, minimum, sqrt, log, abs, negative, inverse, sine, cosine, and tangent, with variables or nested expressions as operands. No more than five operators are needed, every prediction must be within $10^{-6}$ of \texttt{Y}, and computation should be minimized.

\noindent\textbf{Trajectory prefix (excerpt).}
\texttt{[14]} \texttt{\$ python: correlate candidate terms with Y} \\
\hspace*{2em}\texttt{Correlation between 1/X4 and Y: 0.9999994210502484} \quad \texttt{Mean absolute difference for 1/X4: 0.8401576972014005} \\
\texttt{[15]} There is a very strong relationship between \texttt{1/X4} and \texttt{Y}. The mean squared error and the mean absolute difference are small but not zero, which might indicate a small constant term or some other slight adjustment. \\
\texttt{[17]} \texttt{\$ python: examine Y - 1/X4} \\
\hspace*{2em}\texttt{Statistics of Y - 1/X4:} \quad \texttt{mean 0.840158} \quad \texttt{min 0.541056} \quad \texttt{max 1.000000} \quad [...] \\
\hspace*{2em}\texttt{Maximum absolute difference for Y = 1/X4 + 0.54090000: 0.45909996913165596} \\
\texttt{[18]} The difference between \texttt{Y} and \texttt{1/X4} has a strong negative correlation with \texttt{X3} (\texttt{-0.9735}), which suggests \texttt{X3} is involved in the formula. When I tried various constants to add to \texttt{1/X4}, none of them produced a match within the tolerance. I should now test formulas that combine \texttt{1/X4} with \texttt{X3} in various ways.

\noindent\textbf{Candidate A.} Keep \texttt{inverse(X4)} as the base term, then test additive residuals formed by applying each allowed unary operator to \texttt{X3} and compare them by maximum absolute error. Validate the strongest candidate on every row.

\noindent\textbf{Candidate B.} Keep \texttt{inverse(X4)} as the base term, then test additive residuals formed directly from \texttt{X3}, using linear regression to guide simple add and subtract variants and comparing them by maximum absolute error. Validate the strongest candidate on every row.

\noindent\textbf{Supported candidate.} A. In the recorded run, the agent takes B and fits \texttt{Y - 1/X4 = -0.46926192 * X3 + 1.07467532}, and the closest expressible variant \texttt{1/X4 + (1 - X3)} reports \texttt{Maximum absolute error: 5.4016000766e-01}. The agent then applies the allowed unary operators to \texttt{X3}, finds \texttt{add(inverse(X4), cosine(X3))} with a maximum absolute error of \texttt{5.13e-10}, and confirms \texttt{All points within tolerance: True} on all 1000 rows.
\end{exampleitem}


%% file: tables/model_results_full.tex

\begin{tabular}{lrrrrrrrrrrrrr}
\toprule
Model & Average & Eng. & Res. & P-Eng & P-Res & D-Eng & D-Res & Mean acc. & CC & CW & WC & WW & Unparsed \\
\midrule
GPT-5.6 Sol & 59.7 & 56.9 & 62.5 & 75.8 & 56.2 & 48.1 & 67.2 & 65.2 & 292 & 25 & 29 & 156 & 0 \\
GPT-5.5 & 59.5 & 55.6 & 63.4 & 73.4 & 56.2 & 47.4 & 68.8 & 64.6 & 288 & 26 & 24 & 164 & 0 \\
Claude Opus 5 & 55.5 & 46.7 & 64.3 & 71.0 & 64.6 & 35.3 & 64.1 & 60.3 & 254 & 26 & 25 & 197 & 0 \\
Grok 4.5 & 54.6 & 52.1 & 57.1 & 61.3 & 56.2 & 47.7 & 57.8 & 62.0 & 267 & 40 & 43 & 152 & 10 \\
GPT-5.6 Terra & 54.0 & 50.0 & 58.0 & 70.2 & 58.3 & 40.6 & 57.8 & 62.2 & 260 & 40 & 46 & 156 & 0 \\
GLM-5.2 & 53.9 & 47.9 & 59.8 & 64.5 & 60.4 & 40.2 & 59.4 & 64.0 & 254 & 52 & 49 & 147 & 17 \\
Claude Sonnet 5 & 51.6 & 44.4 & 58.9 & 62.1 & 56.2 & 36.1 & 60.9 & 60.7 & 239 & 37 & 52 & 174 & 0 \\
GPT-5.6 Luna & 49.0 & 43.6 & 54.5 & 67.7 & 50.0 & 32.3 & 57.8 & 58.9 & 231 & 48 & 51 & 172 & 0 \\
MiniMax M3 & 45.3 & 44.1 & 46.4 & 64.5 & 56.2 & 34.6 & 39.1 & 57.2 & 224 & 61 & 51 & 166 & 3 \\
DeepSeek V4 Flash & 43.3 & 38.5 & 48.2 & 58.1 & 50.0 & 29.3 & 46.9 & 54.9 & 204 & 59 & 55 & 184 & 2 \\
GPT-5.4 Mini & 40.1 & 25.6 & 54.5 & 26.6 & 54.2 & 25.2 & 54.7 & 49.5 & 161 & 46 & 59 & 236 & 112 \\
Mistral Medium 3.5 & 37.7 & 41.5 & 33.9 & 43.5 & 41.7 & 40.6 & 28.1 & 51.8 & 200 & 70 & 67 & 165 & 24 \\
GPT-5.4 Nano & 36.6 & 32.1 & 41.1 & 46.0 & 43.8 & 25.6 & 39.1 & 48.8 & 171 & 65 & 61 & 205 & 93 \\
Grok 4.20 Reasoning & 15.7 & 22.6 & 8.9 & 28.2 & 8.3 & 19.9 & 9.4 & 29.0 & 98 & 78 & 72 & 254 & 459 \\
\bottomrule
\end{tabular}

%% file: tables/seh_by_model.tex

\begin{tabular}{lrrrr}
\toprule
Model & In prefix & Inferable & Next step & More work \\
& ($n=158$) & ($n=219$) & ($n=56$) & ($n=69$) \\
\midrule
GPT-5.5 & 75.9 & 57.1 & 48.2 & 23.2 \\
GPT-5.6 Sol & 79.7 & 57.5 & 44.6 & 21.7 \\
Grok 4.5 & 67.1 & 53.9 & 32.1 & 36.2 \\
Claude Opus 5 & 67.1 & 47.0 & 50.0 & 24.6 \\
GPT-5.6 Terra & 73.4 & 48.9 & 41.1 & 20.3 \\
GLM-5.2 & 69.0 & 49.8 & 32.1 & 26.1 \\
Claude Sonnet 5 & 63.9 & 45.7 & 35.7 & 26.1 \\
GPT-5.6 Luna & 67.7 & 43.8 & 32.1 & 14.5 \\
MiniMax M3 & 64.6 & 41.1 & 25.0 & 26.1 \\
DeepSeek V4 Flash & 61.4 & 35.6 & 30.4 & 17.4 \\
Mistral Medium 3.5 & 54.4 & 42.5 & 14.3 & 18.8 \\
GPT-5.4 Nano & 49.4 & 31.1 & 25.0 & 15.9 \\
GPT-5.4 Mini & 46.2 & 30.1 & 16.1 & 18.8 \\
Grok 4.20 Reasoning & 32.3 & 16.4 & 14.3 & 4.3 \\
\midrule
Mean over models & 62.3 & 42.9 & 31.5 & 21.0 \\
\bottomrule
\end{tabular}

%% file: tables/effort_by_level.tex

\begin{tabular}{llrrrrr}
\toprule
Model & Setting & Accuracy & In prefix & Inferable & Next step & More work \\
\midrule
GPT-5.6 Sol & low & 56.2 & 79.1 & 53.9 & 44.6 & 20.3 \\
GPT-5.6 Sol & xhigh & 55.6 & 78.5 & 52.1 & 46.4 & 21.7 \\
GPT-5.6 Sol & max & 56.0 & 79.7 & 52.5 & 42.9 & 23.2 \\
\midrule
GPT-5.6 Luna & none & 43.4 & 60.8 & 44.3 & 28.6 & 13.0 \\
GPT-5.6 Luna & high & 45.8 & 67.7 & 42.9 & 28.6 & 18.8 \\
GPT-5.6 Luna & max & 45.6 & 65.2 & 43.4 & 35.7 & 15.9 \\
\bottomrule
\end{tabular}

%% file: tables/human_review.tex
\begin{tabular*}{\linewidth}{@{\extracolsep{\fill}}lrr@{}}
\toprule
Reviewer & Retained judgments & Label agreement \\
\midrule
Reviewer 1 & 87 & \textbf{86/87 (98.9\%)} \\
Reviewer 2 & 85 & \textbf{84/85 (98.8\%)} \\
\midrule
Combined & 172 & \textbf{170/172 (98.8\%)} \\
\bottomrule
\end{tabular*}

%% file: tables/e2e_tasks.tex

\begin{tabular}{lrr}
\toprule
Task & Forks & Student correct \\
\midrule
NodeBB (3c85b94) & 1 & 1 \\
ansible (0fd8871) & 1 & 1 \\
ansible (395e5e2) & 4 & 4 \\
ansible (7094849) & 1 & 1 \\
ansible (7765870) & 3 & 2 \\
ansible (be2c376) & 3 & 3 \\
ansible (cd9c4eb) & 4 & 3 \\
ansible (ea04e00) & 7 & 6 \\
ansible (ed6581e) & 1 & 1 \\
element-web (494d9de) & 1 & 0 \\
element-web (b7fea97) & 1 & 1 \\
element-web (ca8b1b0) & 2 & 2 \\
flipt (3ef34d1) & 3 & 1 \\
flipt (b22f5f0) & 1 & 1 \\
vuls (4c04acb) & 2 & 2 \\
vuls (abd8041) & 1 & 1 \\
vuls (dc49646) & 2 & 2 \\
vuls (e52fa8d) & 1 & 1 \\
teleport (6eaaf3a) & 2 & 2 \\
teleport (baeb269) & 1 & 1 \\
openlibrary (08ac40d) & 3 & 3 \\
openlibrary (3c48b4b) & 3 & 1 \\
openlibrary (3f580a5) & 2 & 2 \\
openlibrary (5899980) & 3 & 3 \\
openlibrary (6a117fa) & 1 & 0 \\
openlibrary (9cd47f4) & 3 & 3 \\
openlibrary (b4f7c18) & 3 & 1 \\
openlibrary (e010b2a) & 1 & 1 \\
navidrome (8383527) & 2 & 0 \\
webclients (01b519c) & 3 & 3 \\
webclients (4feccbc) & 4 & 4 \\
qutebrowser (1943fa0) & 1 & 1 \\
qutebrowser (2dd8966) & 2 & 1 \\
qutebrowser (9902914) & 1 & 1 \\
qutebrowser (996487c) & 1 & 1 \\
qutebrowser (a84ecfb) & 5 & 4 \\
qutebrowser (de4a1c1) & 4 & 2 \\
tutanota (09c2776) & 3 & 1 \\
tutanota (1e516e9) & 4 & 4 \\
tutanota (219bc8f) & 3 & 2 \\
tutanota (da4edb7) & 4 & 3 \\
\midrule
Total & 98 & 77 \\
\bottomrule
\end{tabular}